%% file: main.tex
\documentclass[journal]{IEEEtran}
\usepackage{amsmath}
\usepackage{array}
\usepackage{graphicx}
\usepackage{subcaption}
\usepackage{xcolor}
\usepackage{hyperref}

\begin{document}
\title{When Dance Video Archives Challenge Computer Vision}
%
%
%

\author{Philippe Colantoni, \and Rafique Ahmed, \and  Prashant Ghimire, \and Damien Muselet, \and  Alain Trémeau\\
Laboratoire Hubert Curien - UMR 5516\\
philippe.colantoni@univ-st-etienne.fr, rafique.ahmed@univ-st-etienne.fr, prashant.ghimire@univ-st-etienne.fr, damien.muselet@univ-st-etienne.fr, alain.tremeau@univ-st-etienne.fr
}
%

\markboth{}%
{Shell \MakeLowercase{\textit{et al.}}: Bare Demo of IEEEtran.cls for IEEE Journals}
\maketitle

\begin{abstract}
The accuracy and efficiency of human body pose estimation  depend on the quality of the data to be processed and of the particularities of these data. To demonstrate how dance videos can challenge pose estimation techniques, we proposed a new 3D human body pose estimation pipeline which combined up-to-date techniques and methods that had not been yet used in dance analysis. Second, we performed tests and extensive experimentations from dance video archives, and used visual analytic tools to evaluate the impact of several data parameters on human body pose. Our results are publicly available for research at \url{https://www.couleur.org/articles/arXiv-1-2025/}
\end{abstract}

\section{Introduction}
\input{text/Introduction}
\section{Materials and Methods}\label{section2}
\input{text/MaterialsandMethods}
\section{Results} \label{section3}
Here we present the visual analytical tools which have developed to help a user the select the best result(s) that a pose estimation method can provide depending of the video content.
\input{text/Results}
\section{Discussions} \label{section4}
\input{text/Discussion}
\section*{Funding details}
This work was supported by the HORIZON-CL2-2021-HERITAGE-000201-04 under Grant number 101061303 - PREMIERE \cite{PREMIERE2024}

\section*{Disclosure statement}
 The authors report there are no competing interests to declare.
 
 \section*{Data availability statement}
The two dancers of the “PREMIERE dance motion dataset” are Jeroen Janssen and Rodrigo Ribeiro, they performed at AHK ID-Lab at Amsterdam on 21 Sept. 2023. The PREMIERE dance motion dataset is available on demand.
\bibliography{main}
\bibliographystyle{IEEEtran}
\end{document}

%% file: text/Introduction.tex
\IEEEPARstart{T}{his} paper illustrates some of the digital methods and tools that have been developed in the frame of the PREMIERE project. 

The PREMIERE project aims to modernize performing arts by developing digital tools in order to support the production (for new performances), the curation (for past) of digitized performances, the understanding and the distribution of digitized performances. Digital recordings (video and images) are still rarely unexploited, partly due to the absence of methodological and technical solutions to capture, represent, and exploit the latent information. With the emergence of new technologies and ever-increasing availability of digitized performances, the methodology followed to semantically enrich and utilize such resources becomes a vital factor in supporting professionals’ and users' needs. However, little progress has been made on applying and understanding the role of new technologies in the preservation, transmission and creation of performing arts in general, and specifically dance. Joshi et al. identified in 2021 six major categories in dance automation: dance representation, dance capturing, dance semantics, dance generation, dance processing approaches, and applications of dance automation systems \cite{joshi2021extensive}. The application of motion capture technology in dance was until recently limited to data collection and preservation of performances, and it rarely directly involved dance creation. The most recent progresses in the field are reviewed in \cite{joshi2021extensive} and \cite{reshma2023cultural}, meanwhile \cite{stacchio2024danxe} proposed in a framework based on Artificial Intelligence and Extended Reality for the digitization, automatic analysis, and immersive manipulations of Dance Heritage.

The most accurate way to capture human body movement in a live setting is to use Inertia-based motion capture equipment (XSens, Rokoko, etc.). However, with marker-based motion capture systems, performers are wearing suits which is problematic in the case of dance performances, as suits may restrict natural movement and may cause discomfort for performers. Some marker-based motion capture systems are equipped with infrared sensors which limits the use of certain lighting conditions, which is problematic for some dance performances. Moreover, these systems involve extensive setup and calibration, which can be time-consuming and require technical expertise. Until recently, these systems were more accurate and more expensive than markerless motion capture systems.  However, advancements in AI and machine learning have rapidly closing this gap. These systems do not require use of complex equipment reducing setup complexity, they are easier to use in various environments without the need for a controlled studio setup. Moreover, performers can move freely and naturally, resulting in more authentic motion data. Camera sensors and video-based methods are less intrusive but require more processing and have problems with accurate 3D localization. Several works have been done to evaluate and compare the accuracy, the efficiency, and the advantage and drawback, of marker-based motion capture systems versus markerless motion capture systems, to cite a few:  Nobuyasu et al. made an evaluation of the accuracy of a 3D markerless motion capture system with multiple video cameras using OpenPose \cite{nakano2020evaluation}; Scataglini et al. studied the accuracy, validity, and reliability of markerless camera-based 3D motion capture systems versus marker-based 3D motion capture systems \cite{scataglini2024accuracy}. In our tests and experiments, we used GoPro cameras to capture the performers’ motion, with a minimal, non-intrusive, and affordable setup. We also conducted tests and experiments with archive videos provided by PREMIERE partners, some of them are illustrated below.

The automatic semantic analysis and enrichment of performing arts data is considered as one of the most fundamental challenge to solve, especially in dance, since performers add complexities related to events to the classical human body descriptors associated with physical objects. The most relevant lacks of the present models are related to the modeling of information connected to performers and performances, which are obviously distinguishing aspects of the performing arts. The objective of this paper is to push the state-of-the-art in digital methodological and technical solutions in order to automatically semantically enrich digitized performances. To fulfill this objective, we investigated scene understanding and 3D reconstruction methods. The most commonly used methods are based on computer vision and machine learning methods to perform different tasks, such as pose and motion estimation, body pose trajectories estimation, tracking, and action parsing from video recordings. However, the up-to-date methods are still not efficient enough to accurately analyze and reconstruct a 3D scene, or to recognize a complex and similar dance action \cite{qianwen2024application}. In our investigations, we studied how to expand these methods to precisely estimate the position of the performers' bodies on a stage from videos and from this estimate set up a complete analysis of their different movements/interactions by relying on several works that have already been done in this field by tackling inter-object occlusions, abrupt motion changes, appearance changes due to different lighting conditions, missing detections due to non-rigid deformation of clothing, truncations of persons, or interactions between people.

The accuracy and efficiency of a video processing pipeline depend first of the quality of the video files to be processed. Four main criteria have to be considered to assess the quality of a video \cite{PREMIERE2023}.
\begin{itemize}
\item The first one concerns the acquisition quality, it is related to the acquisition devices and their settings. The main points that should be checked before acquisition are the frame-rate and the resolution. A low frame-rate (less than 25 FPS), compared to motion speed, can lead to strong motion blur that disturbs many image processing algorithms (as example see Figure \ref{fig1}), while low resolution does not allow to detect small objects or to estimate face emotion.
\begin{figure}[h!]
\centering
\includegraphics[width=0.175\textwidth]{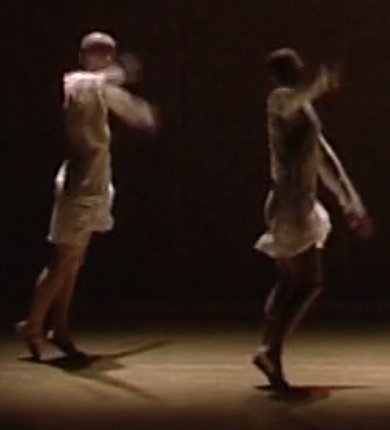}
\includegraphics[width=0.13\textwidth]{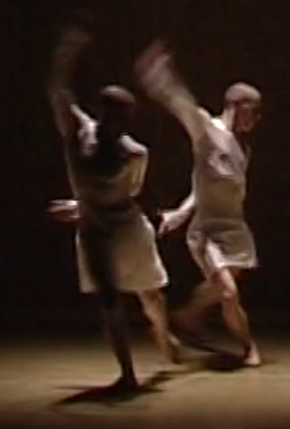}
\caption{Motion blur on left and right images. Images from PREMIERE video archive ``Extra Dr'' from Emio Greco - 1999 (resolution 1050 x 578).}
\label{fig1}
\end{figure}

\item The second criterion is about the lighting conditions. Many archive videos used in the PREMIERE project were acquired under low light conditions that increase the image noise and the motion blur. Also, light spatial variations are not desirable because they cause under- and over-exposed areas in the images. In the context of the PREMIERE project, some archive videos have lots of light shows included in the performance. These specific lighting conditions are difficult to handle by generic algorithms (as example, see Figures \ref{fig2} and \ref{fig3}).

\begin{figure}[ht!]
    \centering
    \begin{minipage}{0.22\textwidth}
        \centering
        \includegraphics[width=\textwidth]{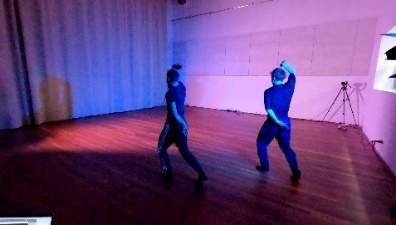}
        \subcaption{}
        \label{fig2.1}
    \end{minipage}%
    \hspace{0cm}
    \begin{minipage}{0.22\textwidth}
        \centering
        \includegraphics[width=\textwidth]{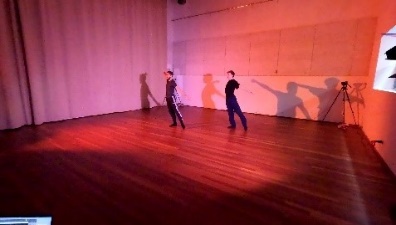}
        \subcaption{}
        \label{fig2.2}
    \end{minipage}
    \vskip\baselineskip
    \begin{minipage}{0.22\textwidth}
        \centering
        \includegraphics[width=\textwidth]{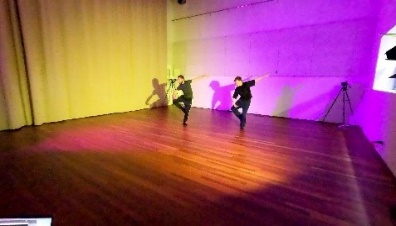}
        \subcaption{}
        \label{fig2.3}
    \end{minipage}
    \caption{Scene lighted with two color spots. (a) dancers are in the shadow area, (b) color shifts induced by changes of lighting color over time, (c) complex shading effects induced by changes of lighting color over time. Images from the PREMIERE Dance Motion Dataset.}
    \label{fig2}
\end{figure}

\begin{figure}[ht!]
    \centering
    \begin{minipage}{0.22\textwidth}
        \centering
        \includegraphics[width=\textwidth, keepaspectratio]{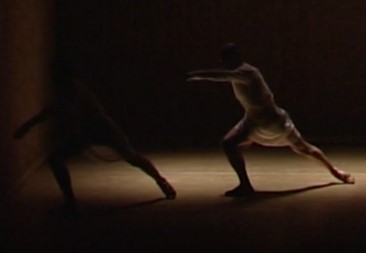}
        \subcaption{}
        \label{fig3.1}
    \end{minipage}%
    \hspace{0cm}
    \begin{minipage}{0.22\textwidth}
        \centering
        \includegraphics[width=\textwidth, keepaspectratio]{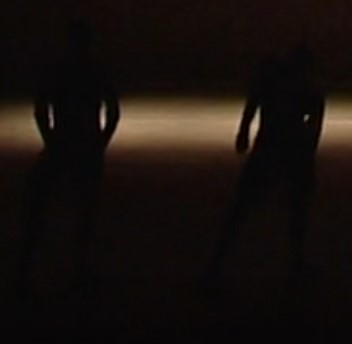}
        \subcaption{}
        \label{fig3.2}
    \end{minipage}
    \vspace{0cm}
    \begin{minipage}{0.22\textwidth}
        \centering
        \includegraphics[width=\textwidth, keepaspectratio]{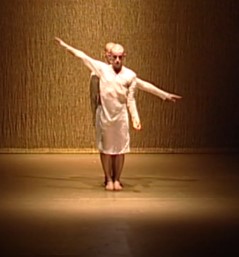}
        \subcaption{}
        \label{fig3.3}
    \end{minipage}
    \hspace{0cm}
    \begin{minipage}{0.22\textwidth}
        \centering
        \includegraphics[width=\textwidth, keepaspectratio]{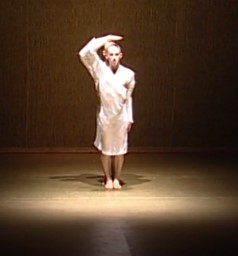}
        \subcaption{}
        \label{fig3.4}
    \end{minipage}
    \caption{(a) and (b) Very dark lighting, Low/non-uniform color contrast between dancers and the background and Low color contrast between dancers' feet and the floor, due to shadow. (c) and (d) Color variations of light. Images from PREMIERE video archive ``Extra Dry'' from Emio Greco - 1999 (resolution 1050 x 578).}
    \label{fig3}
\end{figure}

\item The third criterion that can be interesting to check in a performance video is the relative positioning of the camera, the light sources and the performers. For some videos, we observed important close-ups of the bodies or faces of the performers, making the automatic interpretations very difficult. Furthermore, these relative positions can create shadows on the background and fool the detection of human or objects.
\item Lastly, the fourth important criterion is the scene and the performance themselves. For example, when occlusions or body contacts are occurring, the human pose detection can be disturbed (as example see Figures \ref{fig4} and \ref{fig5}). Likewise, loose clothing can be a very hard challenge in this context (as example, see Figure \ref{fig6}). There also can be some performances where the audience is visible and this provides some additional data to process, even if the extracted features considered are not interesting for the analysis of the performance. 

\begin{figure}[ht!]
    \centering
    \begin{minipage}{0.22\textwidth }
        \centering
        \includegraphics[width=\textwidth,keepaspectratio]{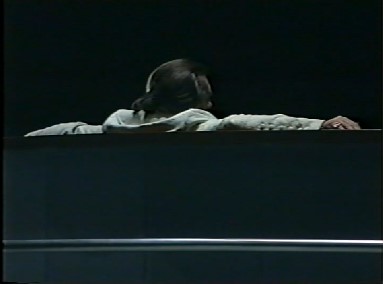}
        \subcaption{}
        \label{fig4.1}
    \end{minipage}%
    \hspace{0cm}
    \begin{minipage}{0.22\textwidth}
        \centering
        \includegraphics[width=\textwidth,keepaspectratio]{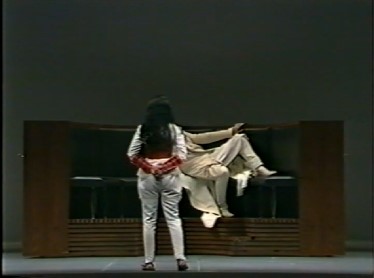}
        \subcaption{}
        \label{fig4.2}
    \end{minipage}
    \caption{ Examples of strong occlusions;  (a) image (2D view), where only the upper part of the human body is visible. (b) image (2D view), where  the upper part of the human body in the background is occluded. The 2D views shown are not optimal for human pose estimation. Images from PREMIERE video archive ``Noite de Reis''  from Ricardo Pais – 1998 (resolution 720x540).}
    \label{fig4}
\end{figure}

\begin{figure}[ht!]
    \centering
    \begin{minipage}{0.22\textwidth}
        \centering
        \includegraphics[width=\textwidth]{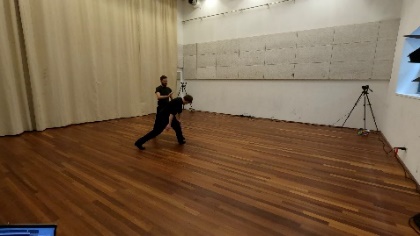}
        \subcaption{}
        \label{fig5.1}
    \end{minipage}%
    \hspace{0cm}
    \begin{minipage}{0.22\textwidth}
        \centering
        \includegraphics[width=\textwidth]{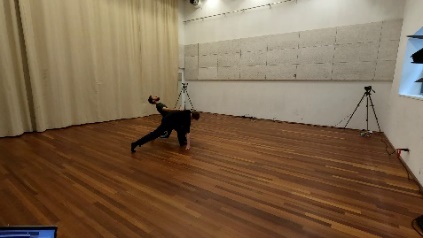}
        \subcaption{}
        \label{fig5.2}
    \end{minipage}
    \vspace{-0.3cm} 
    \begin{minipage}{0.22\textwidth}
        \centering
        \includegraphics[width=\textwidth]{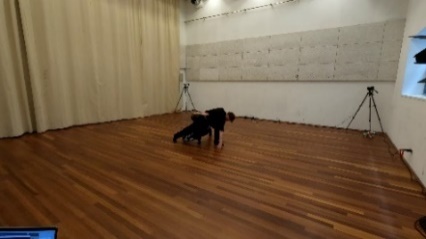}
        \subcaption{}
        \label{fig5.3}
    \end{minipage}
    \caption{Various occlusion examples: (a) one of the dancers occludes the lower part of the body of the second dancer, (b) complex occlusion pattern, (c) another challenging occlusion pattern. Images the PREMIERE Dance Motion Dataset.}
    \label{fig5}
\end{figure}

\begin{figure}[ht!]
    \centering
    \begin{minipage}{0.22\textwidth}
        \centering
        \includegraphics[width=\textwidth, keepaspectratio ]{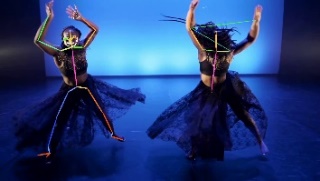}
        \subcaption{}
        \label{fig6.1}
    \end{minipage}%
    \hspace{0cm}
    \begin{minipage}{0.22\textwidth}
        \centering
        \includegraphics[width=\textwidth, keepaspectratio]{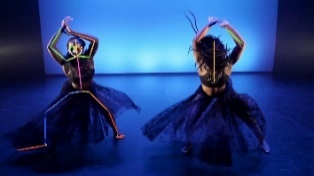}
        \subcaption{}
        \label{fig6.2}
    \end{minipage}
   \vspace{-0.1cm}
    \begin{minipage}{0.22\textwidth}
        \centering
        \includegraphics[width=\textwidth, keepaspectratio]{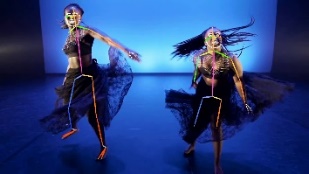}
        \subcaption{}
        \label{fig6.3}
    \end{minipage}
    \hspace{0cm}
    \begin{minipage}{0.22\textwidth}
        \centering
        \includegraphics[width=\textwidth, keepaspectratio]{images/Introduction/fig6_3.jpg}
        \subcaption{}
        \label{fig6.4}
    \end{minipage}
    \caption{Examples of wrong 2D pose estimation due to the loose clothing of the dancers and low light conditions. For example in (b) legs pose estimation of the dancer on right is wrong because of the cloth. All 4 image sequence (a), (b), (c) and (d) comes from \cite{Talawa2018}.}
    \label{fig6}
\end{figure}
\end{itemize}

To face these challenges, we propose in section \ref{section2} a new methodology for 3D human body pose estimation and scene reconstruction. The new video processing pipeline described in this paper combines up-to-date techniques and methods that had not been yet used in dance analysis, they were mainly used for other body motion activities, such as sports. None of the methods used has been cited in the survey on dance digitalization of \cite{reshma2023cultural} published in 2023 for example.

Visual analytic tools can be used in order to help a user to: identify the gesture, poses, and stance; recognize the dance forms, and compare complex and similar dance action; classify dance movements; transform 3D human model of dance performers in 3D avatars, etc. \cite{guo2021phycovis, arpatzoglou2021dancemoves, reshma2023cultural, zhang2023neuromorphic}. As example, Zhang et al. implemented customized scripts and inject them into Blender to: collect the exact position of all human’ joints while rendering scenes at 300 FPS; extract the camera’s intrinsic and extrinsic matrix to accurately estimate the 3D position in the space of the dancers; and create more realistic 3D human models \cite{zhang2023neuromorphic}. Lee et al. proposed in \cite{lee2024beyond} a dancer viewer that reconstructs dancers in video into 3D avatars and a dance feedback tool that analyzes and compares the user’s performance with that of the reference dancer. Visual analytic tools can be also used to help a user to: select the best pose estimation method(s) depending on the lighting conditions for example, adjust some parameters of pose estimation algorithms to refine the accuracy of the results, refine the computation of body trajectories in case of body occlusions for example, etc.

In section \ref{section3}, we introduce new visual analytical tools in order to help the user to: - select the information needed; - assess the impact of some parameters on human body pose and trajectories estimation; - refine the results obtained; - display 3D human model in augmented reality (AR) or eXtended Reality (XR) modes. These methods and tools can be used to boost understanding of dance’s underlying semantics.

Lastly, in section \ref{section4} we draw a conclusion, discuss the limitations of the tools implemented, and suggest future research directions..

%% file: text/MaterialsandMethods.tex
\subsection{Introduction}
The video processing pipeline described in the following sections was developed to estimate and track the movements of multiple dancers from a single video. The complexity of the videos we had to process for the PREMIERE project led us to develop multiple processing pipelines adapted to the content of the video sequences. In this paper, we will present the processing pipeline adapted to video sequences with a fixed number of dancers, all of whom are present in the sequence at a given time, and to slight camera movements. We chose this scenario because it was considered particularly relevant by the choreographers and dancers who are members of the PREMIERE project. Other scenarios in which dancers appear, disappear, and reappear in a video sequence are handled by a another pipeline which will be not discussed in this paper.

\begin{figure}[ht!]
\centering
\includegraphics[width=0.45\textwidth]{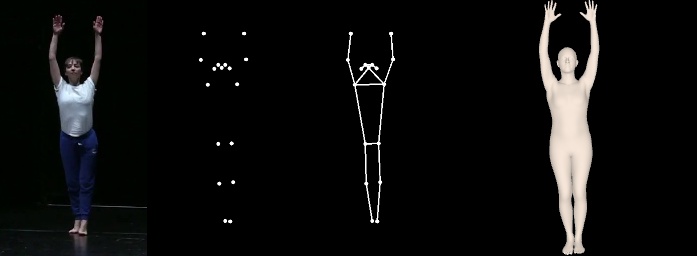}
\caption{(From left to right) RGB image, major joints, skeleton model, SMPL representation.}
    \label{fig7}
\end{figure}

In addition to 3D pose estimation of dancers, this pipeline can also extract their SMPL-X representations. SMPL-X \cite{pavlakos2019expressive} is an advanced 3D human body model that extends the original SMPL (Skinned Multi-Person Linear) model by incorporating not only the body but also expressive facial features and articulated hands (see Figure \ref{fig7}). This model captures human shape and pose using a parametric representation built on a kinematic tree structure. In addition to body pose, SMPL-X introduces facial expressions and hand gestures, which are modeled separately. The face is represented using FLAME (Facial Model) parameters \cite{zheng2023flame}, while the hands use a MANO-based hand model with joint rotations \cite{xie2024ms}. This enables the model to capture intricate facial expressions and hand articulations, allowing for a more expressive and realistic representation of human movement and interaction.

The SMPL model and its derivatives offer powerful capabilities in the domain of performing arts. Originally designed for computer graphics and computer vision applications, particularly for 3D human pose and shape estimation from 2D images and videos, they can be used to adapt 3D models to archived videos. Additionally, they excel at creating realistic 3D character animations, which are invaluable for bringing virtual characters to life on stage or in digital productions in dance. Enriched with motion capture data, these models can accurately describe the movements of dancers, facilitating choreography development, rehearsals, and precise movement analysis. Additionally, they can assist costume designers by estimating performers' body shapes and dimensions, allowing for the creation of tailored outfits for improved aesthetics. In dance auditions, these models can also provide preview tools to help directors make casting decisions.

In dance performances, MANO finds it use in recognizing and interpreting complex hand gestures and movements, providing insights into the dancers' expressive choreography. It can seamlessly integrate with lighting and effects systems to facilitate gesture-controlled adjustments to stage elements, injecting dynamism into performances. For injury rehabilitation or therapeutic dance programs, MANO can be used to track and analyze hand movements. In addition, it can also facilitate hand-based interactions between performers and digital elements, crafting immersive and interactive dance experiences.

The FLAME facial model is ideal for creating realistic facial animations. In performing arts, such as theater, it can be used to enhance storytelling and character representation by animating digital characters or avatars. Its facial modelling capabilities can be extended to makeup and costume artists to help them create custom makeup and prosthetics that precisely match actors' facial features, ensuring seamless character integration. The integration of FLAME with facial tracking technologies enable to capture and analyse actors' facial expressions during live performances, it also enables real-time adjustments to digital characters or lighting effects.

In the field of performing arts, SMPL-X, together with FLAME, enable to create virtual avatars or characters that mirror the movements and expressions of real dancers. This is particularly useful for immersive experiences and remote performances. The combination of SMPL-X, FLAME, and MANO can be used as a teaching tool in dance education, improving students' understanding of body movement and anatomy, thereby refining their performance skills. In modern dance productions that frequently use digital effects and projections, these 3D models can help to create crafting realistic digital effects, providing accurate representations of human bodies for interaction with virtual elements. These models integrate perfectly into interactive and augmented reality performances, where the performers' movements are tracked to control the digital elements in real-time.

The focus of this paper in on 3D human-body model based on SMPL-X representation (see \cite{PREMIERE2024}), nevertheless the video processing pipeline introduced in the following section could be extended to more detailed human-body representations such as FLAME and MANO. 
\subsection{Video processing pipeline}

The video processing pipeline proposed here uses recent models of the state of the art and custom algorithms. The steps of this pipeline are:
\begin{itemize}
\item [] Step 1: Archive video cut detection.
\item [] Step 2: Recovering 3D SMPL-X human poses.
\item [] Step 3: Ghost detection removal based on clustering. 
\item [] Step 4: Estimation of position and orientation of cameras.
\item [] Step 5: 3D tracking based on skeleton distances.
\item [] Step 6: Segmentation, tracking and re-identification of people.
\item [] Step 7: Re-identification cleaning.
\item [] Step 8: Tracks filtering and interpolation.
\item [] Step 9: Estimation of camera motion and 3D reconstruction.
\end{itemize}

This pipeline is compatible with any video format, where the performers are on a stage with sufficient pixel density to differentiate them. High-quality images are essential for optimal results.
\subsubsection{Step 1: Archive video cut detection} 
\begin{figure}[ht!]
\centering
\includegraphics[width=0.45\textwidth]{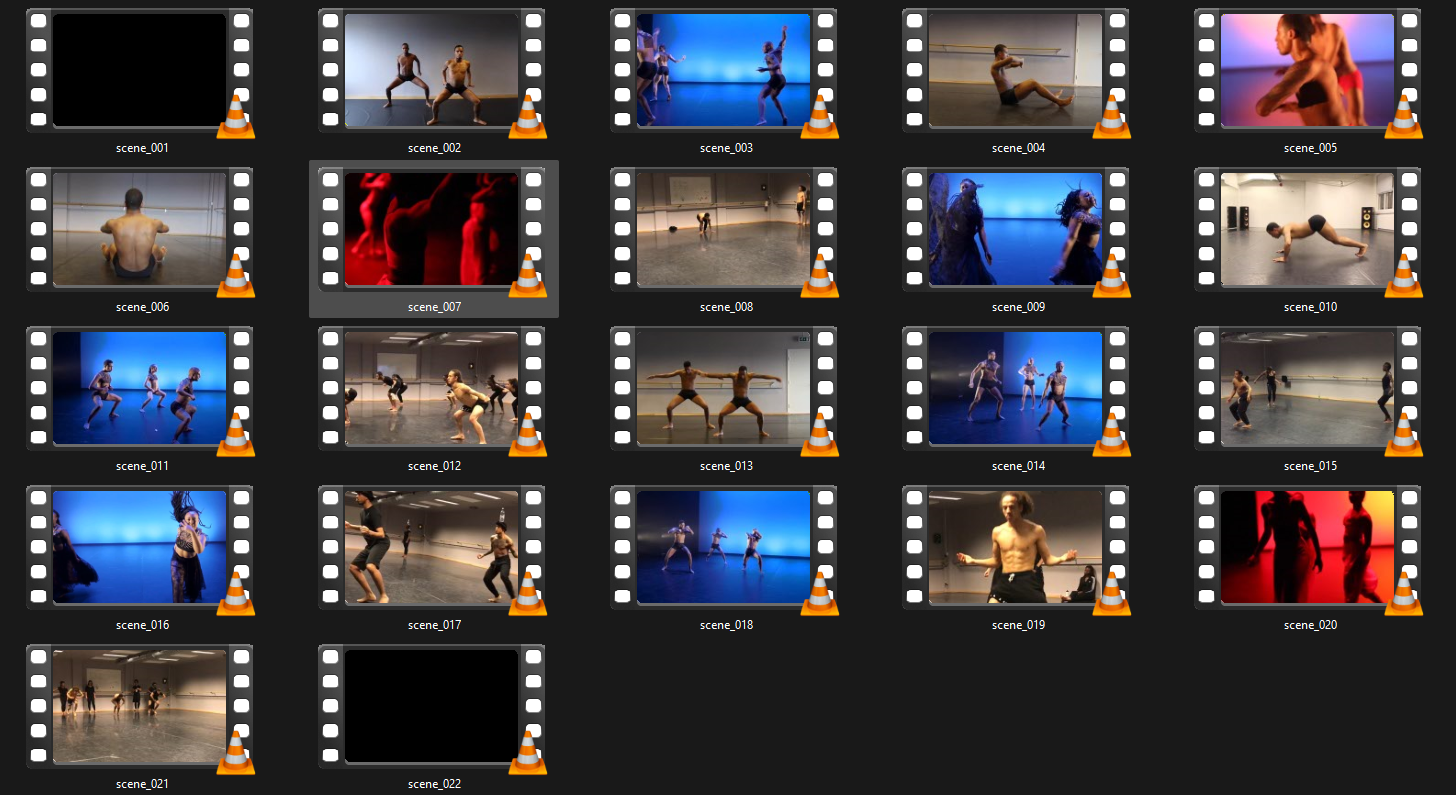}
\caption{Individual sequences detected in the Talawa video. Image comes from \cite{Talawa2018}.}
    \label{fig8}
\end{figure}

The first task consists to automatically cut the video into individual scenes (i.e. to detect scene content changes) as shown in Figure \ref{fig8}. It exists in the State-Of-the-Art efficient methods to detect shot changes in videos, such as PySceneDetect\footnote{https://github.com/Breakthrough/PySceneDetect}.


\subsubsection{Step 2: Recovering 3D SMPL-X human poses} \label{2.2.2} 
To recover the whole human body by a 3D mesh, we used Multi-HMR - Multi-Person Whole-Body Human Mesh Recovery in a Single Shot \cite{baradel2025multi}. Multi-HMR is a single-shot model introduced in February 2024 for recovering 3D human meshes from a single RGB image. It can predict not only the whole-body pose of multiple people in the image but also their shape and 3D location. Multi-HMR can use camera intrinsics to improve performance. Multi-HMR uses a standard Vision Transformer (ViT) backbone \cite{dosovitskiy2020image} and a new Human Prediction Head (HPH) model (as cross-attention module) to make these predictions. Multi-HMR was trained on a new dataset called CUFFS, which contains close-up frames of full-body subjects with diverse hand poses. 

\begin{figure}[ht!]
    \centering
    \begin{minipage}{0.22\textwidth}
        \centering
        \includegraphics[width=\textwidth]{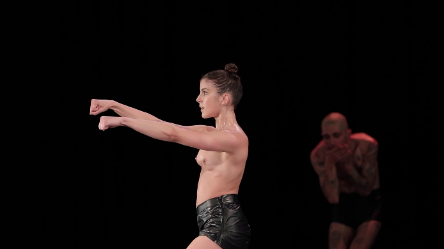}
         \subcaption{}
        \label{fig9.1}
    \end{minipage}%
    \hspace{0cm}
    \begin{minipage}{0.22\textwidth}
        \centering
        \includegraphics[width=\textwidth]{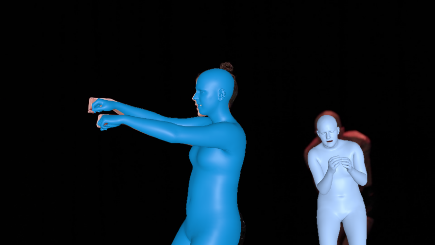}
         \subcaption{}
        \label{fig9.2}
    \end{minipage}
    \vspace{0cm}
    \begin{minipage}{0.22\textwidth}
        \centering
        \includegraphics[width=\textwidth]{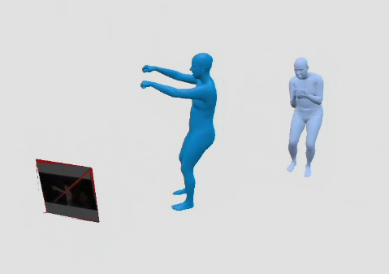}
        \subcaption{}
        \label{fig9.3}
    \end{minipage}
    \caption{(a) RGB image, (b) 3D model superimposed on 2D image, (c) 3D model with camera position in the world coordinate system. Image from PREMIERE video archive “Fecundação e Alívio Neste Chão Irredutível onde Com Gozo Me Insurjo” from Joana von Mayer Trindade \& Hugo Calhim Cristóvão – 2021 (resolution 1920x1080).}
    \label{fig9}
\end{figure}

As illustrated in Figure \ref{fig9}, Multi-HMR enables the extraction of the following data from each frame: 
\begin{enumerate}
    \item 3D keypoints (in the coordinate system of the camera).
    \item 2D keypoints (in the coordinate of the video).
    \item SMPL-X poses.
\end{enumerate}


\begin{figure}[ht!]
\centering
\includegraphics[width=0.35\textwidth]{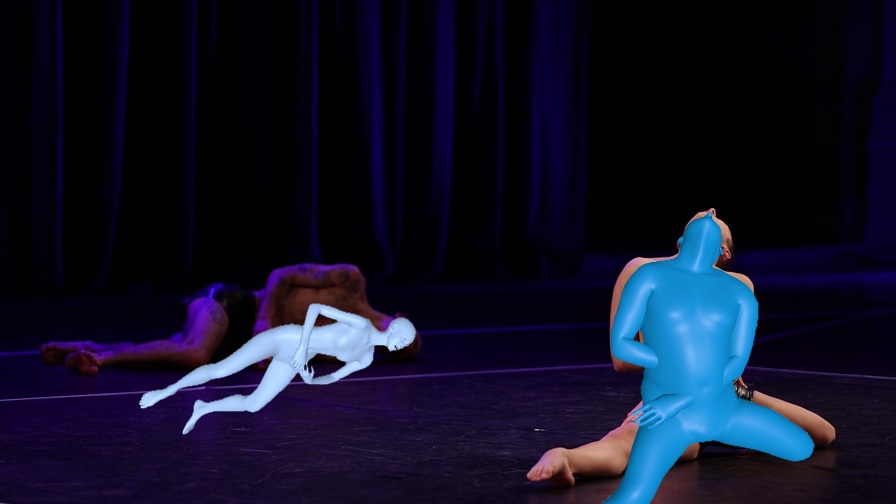}
\caption{3D models superimposed on 2D image: example of inaccuracies.}
    \label{fig10}
\end{figure}

The tests and experiments we have carried out have shown that this model suffers from several shortcomings:
\begin{itemize}
    \item It does not estimate correctly all the dancers' poses (see Figure \ref{fig10}). This problem is partly due to the not-common body poses in the dance which were not used to train this model.
\end{itemize}

\begin{figure}[ht!]
\centering
\includegraphics[width=0.35\textwidth]{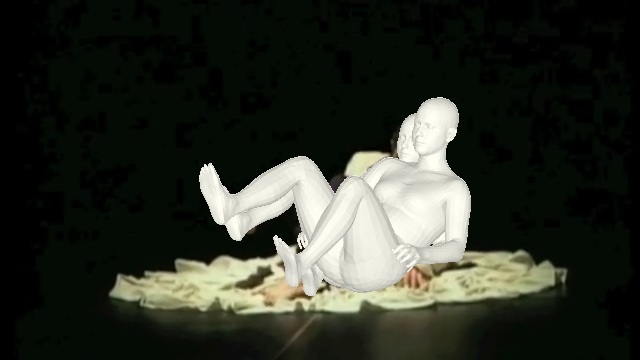}
\caption{3D model superimposed on 2D image generating a ghost pose.}
    \label{fig11}
\end{figure}

\begin{itemize}
    \item It can detect 2 humans when there is only one, as shown in Figure \ref{fig11}.
\end{itemize}

\begin{itemize}
    \item 3D position detection is far from perfect, especially in terms of depth.
    \item The position of the dancers is given in the coordinate system of the camera. We need to determine the extrinsic parameters of this camera to compute the 3D poses in the world coordinate system and not in the camera coordinate system.
\end{itemize}

\subsubsection{Step 3: Ghost detection removal based on clustering}

3D pose estimation models usually depend on external person detection models. Some examples of these models are Detectron2 \cite{wu2019detectron2}, YOLOv7 \cite{wang2023yolov7} or Multi-HMR. It is important to note that other YOLO models can also be utilized. The main advantage of Multi-HMR is that it is based on a single-shot model which can detect humans in a 2D image and their 3D poses during inference. But, this model can sometimes detect several people meanwhile only one is present. This especially happens when lighting conditions are challenging or when two dancers are close to each other. 

\begin{figure}[ht!]
    \centering
    \begin{minipage}{0.22\textwidth }
        \centering
        \includegraphics[width=\textwidth,keepaspectratio]{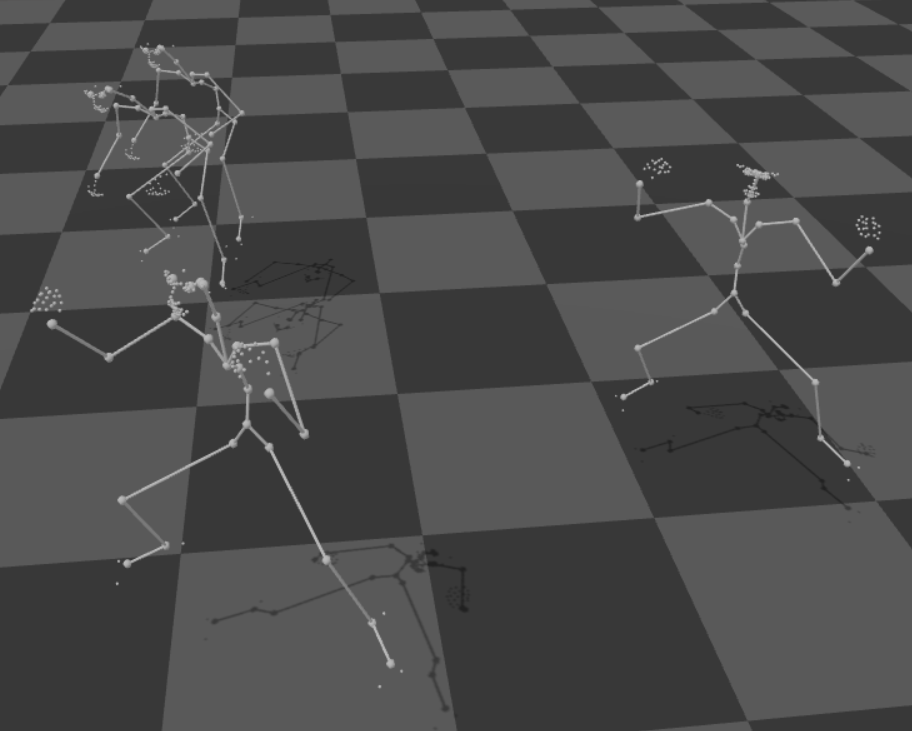}
        \subcaption{}
        \label{fig12.1}
    \end{minipage}%
    \hspace{0cm}
    \begin{minipage}{0.22\textwidth}
        \centering
        \includegraphics[width=\textwidth,keepaspectratio]{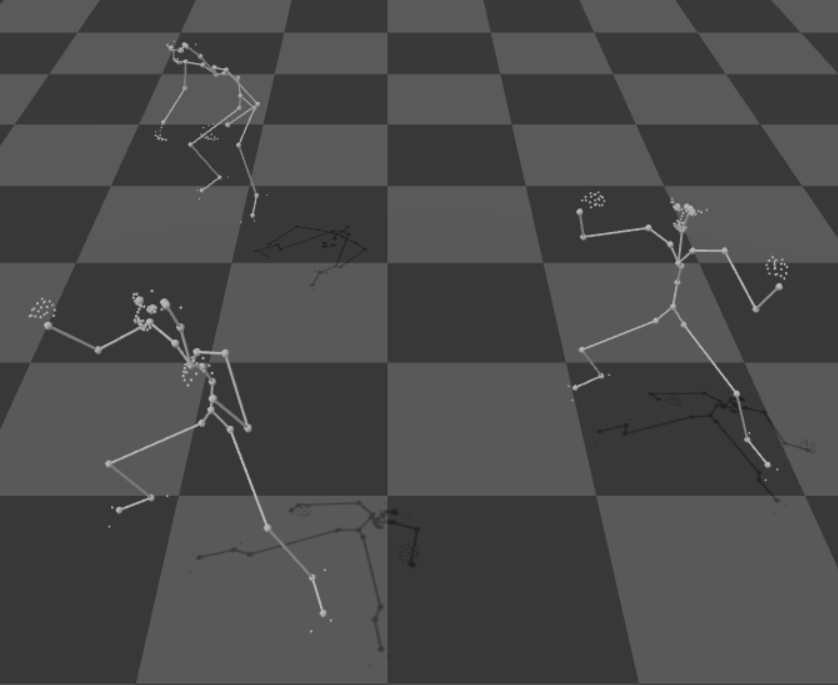}
        \subcaption{}
        \label{fig12.2}
    \end{minipage}
    \caption{ (a) Skeleton before ghost removal (b) Skeleton refinement after ghost removal. As an example, see the dancer in the upper-left area of the image (b), two skeleton models fit the data related to this dancer. After removal of the skeleton model with the lowest confidence score, only one single skeleton model remains. This issue happens for smaller skeletons in the background.}
    \label{fig12}
\end{figure}

It is possible to clean all ghost poses generated by Multi-HMR (as example see result shown in Figure \ref{fig12}) by using a clustering technique. The technique we used is based on a hierarchical clustering on a set of points (the pelvis position of the skeleton) using Ward's method \cite{ward1963hierarchical} and then determines clusters based on a distance threshold. We consider that the pelvis points of two skeletons cannot be less than 40 cm apart. This value has been defined empirically from tests and experiments done with dance video archives related to the PREMIERE project. 
\subsubsection{Step 4: 3D tracking based on skeleton distances} 
\begin{figure}[ht!]
    \centering
    \begin{minipage}{0.23\textwidth }
        \centering
        \includegraphics[width=1.1\textwidth,keepaspectratio]{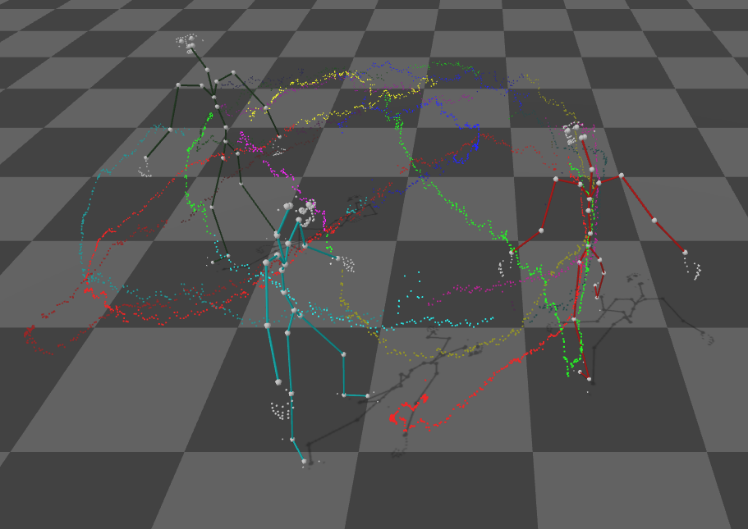}
        \subcaption{}
        \label{fig13.1}
    \end{minipage}%
    \hspace{0cm}
    \begin{minipage}{0.22\textwidth}
        \centering
        \includegraphics[width=0.82\textwidth,keepaspectratio]{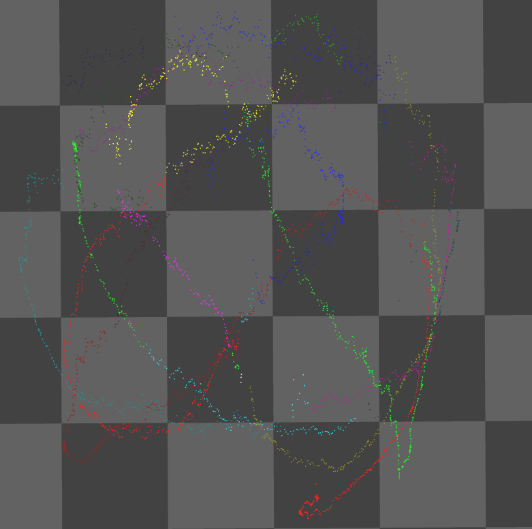}
        \subcaption{}
        \label{fig13.2}
    \end{minipage}
    \caption{ Estimation of trajectories based on tracking of 3D skeletons. Each color corresponds to a track ID. (a)  trajectories and 3D skeletons are shown for one point of view, (b) trajectories are shown from above.}
    \label{fig13}
\end{figure}

We implemented a naive tracking technique based on the tracking of the pelvis point of the skeletons (see Figure \ref{fig13}). It allows us to determine, using a distance threshold, for each skeleton detected if the trajectory of this skeleton is coherent from one frame to the following frame, and if the two skeletons detected in two consecutive frames are part of the same track. This algorithm assigns an ID number for each person present in each frame.

The distance threshold is in meters and depends on the number of frames per second of the video being processed. For a video at 100 FPS, we use a threshold of 30 cm, and for a video at 30 FPS, we use a threshold of 50 cm. These values have been defined empirically from tests and experiments done with dance video archives.

\subsubsection{Step 5: Estimation of position and orientation of cameras} 
To estimate the position and the orientation of cameras (i.e. extrinsic parameters), as well as intrinsic parameters of cameras (e.g. the focal length), we used DUSt3R - Geometric 3D Vision Made Easy \cite{wang2024dust3r}. DUSt3R can reconstruct (from only a few views, or even one image)  2D space into 3D, and extracts geometric information (such as depthmap, 2D view correspondences) in only a few seconds.

The DUSt3R network architecture is based on standard Vision Transformer (ViT) encoders and decoders \cite{wang2024dust3r}. The architecture is inspired by CroCo \cite{weinzaepfel2022croco}, a cross-view completion pre-training pipeline that can understand the spatial relationship between pairs of images. DUSt3R is based on an optimization procedure that globally aligns pointmaps in the context of multi-view 3D reconstruction.

\begin{figure}[ht!]
    \centering
    \begin{minipage}{0.23\textwidth }
        \centering
        \includegraphics[width=\textwidth,keepaspectratio]{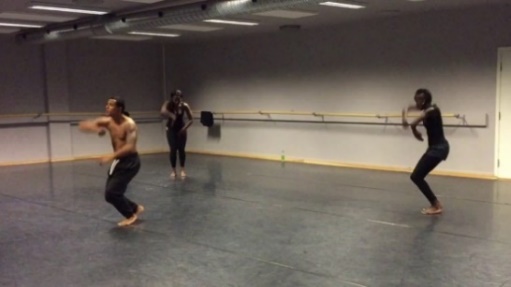}
        \subcaption{}
        \label{fig14.1}
    \end{minipage}%
    \hspace{0cm}
    \begin{minipage}{0.23\textwidth}
        \centering
        \includegraphics[width=\textwidth,keepaspectratio]{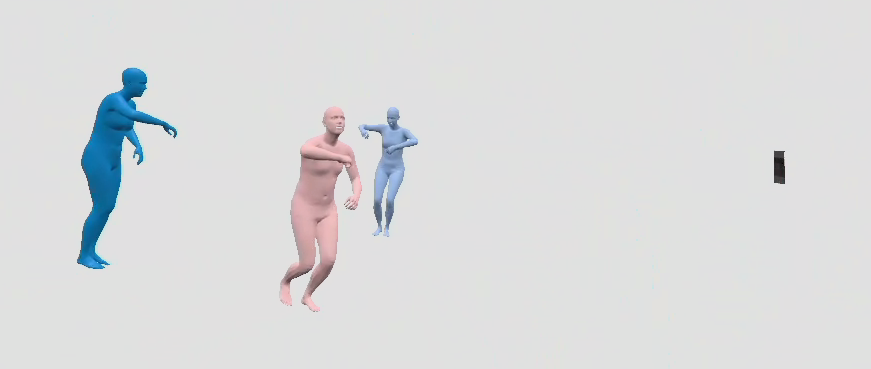}
        \subcaption{}
        \label{fig14.2}
    \end{minipage}
    \caption{ (a) RGB image that comes from \cite{Talawa2018}. (b) 3D model in the world coordinate system from another viewing direction than the camera viewing direction.}
    \label{fig14}
\end{figure}

As mentioned in section \ref{2.2.2}, Multi-HMR generates poses in the coordinate system of the camera used to capture the video. This means that the poses generated from one camera are not in the same plane as the ones generated from another camera, that is why pose estimation must be computed in the world coordinate system (see Figure \ref{fig14}).

\begin{figure}[ht!]
    \centering
    \begin{minipage}{0.24\textwidth }
        \centering
        \includegraphics[width=1.15\textwidth,keepaspectratio]{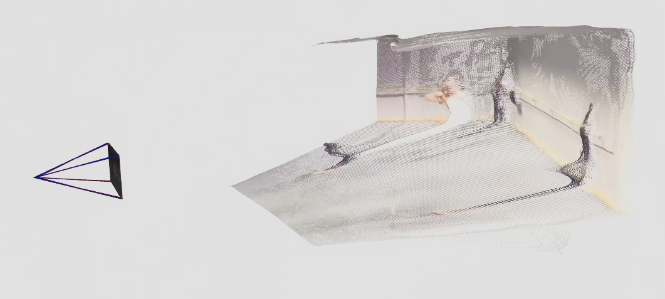}
        \subcaption{}
        \label{fig15.1}
    \end{minipage}%
    \hspace{0cm}
    \begin{minipage}{0.24\textwidth}
        \centering
        \includegraphics[width=.8\textwidth,keepaspectratio]{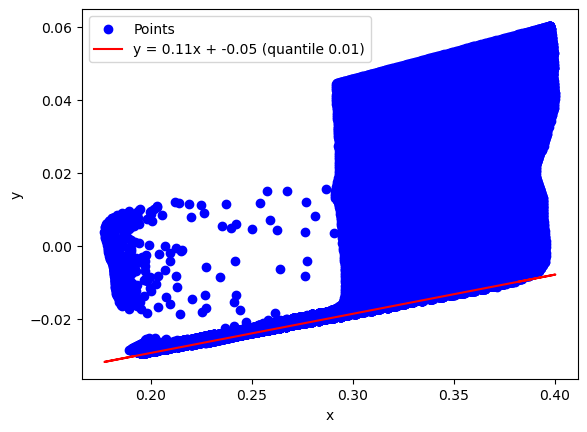}
        \subcaption{}
        \label{fig15.2}
    \end{minipage}
    \caption{ (a) 3D point cloud corresponding to the left image of Figure. (b) 1D fitting model applied to ground data computed from the camera point of view (x represents depth in the world coordinate system and y the height).}
    \label{fig15}
\end{figure}

We use the 3D point cloud generated by DUSt3R. Then, we project all the points onto the (y, z) plane to determine the line corresponding to the stage on which the dancers are moving (see Figure \ref{fig15}). The analytical method used to compute this line is designed to find points that lie below a given line in a 2D space, using a quantile-based approach.

\begin{figure}[ht!]
    \centering
    \begin{minipage}{0.22\textwidth }
        \centering
        \includegraphics[width=\textwidth,keepaspectratio]{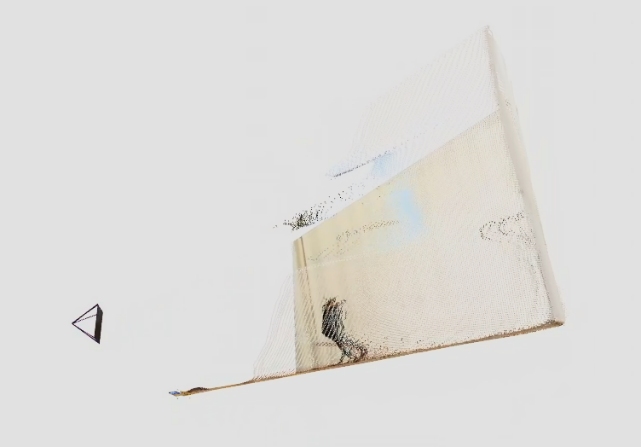}
        \subcaption{}
        \label{fig16.1}
    \end{minipage}%
    \hspace{0cm}
    \begin{minipage}{0.22\textwidth}
        \centering
        \includegraphics[width=\textwidth,keepaspectratio]{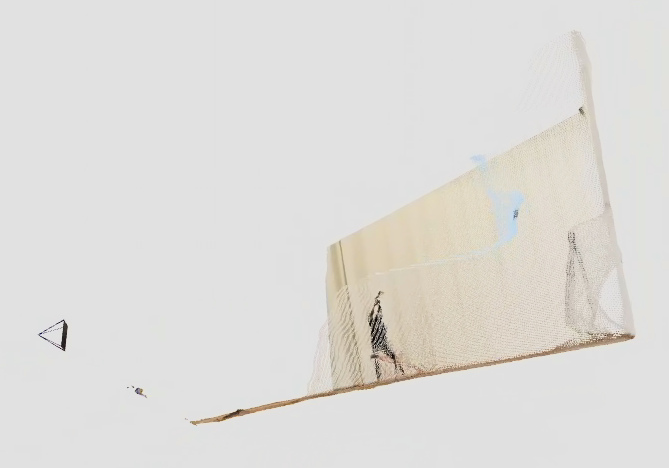}
        \subcaption{}
        \label{fig16.2}
    \end{minipage}
    \caption{ (a) 3D scene reconstructed by DUSt3R (b) 3D scene reconstructed by MASt3R. Note that in (a) the 3D model is flatter than in (b).}
    \label{fig16}
\end{figure}

Additionally, MASt3R can be used to estimate the motion of cameras. MASt3R, which is an evolution of DUSt3R, is said to be more accurate at reconstructing a 3D scene as a point cloud. We conducted a comparative study of these two models and concluded that DUSt3R is more accurate in reconstructing 3D scenes in which the dancers move (as example see Figure \ref{fig16}) than MASt3R.

\subsubsection{Step 6: Segmentation, tracking and re-identification of people} 

To segment human bodies in single RGB images, we used SAM2 - Segment Anything Model 2 \cite{ravi2024sam}. SAM2 is a new real-time image and video segmentation model introduced in August 2024. SAM2 is more accurate than the previous version of SAM model. The previous version of SAM was built explicitly for use in images. SAM2 can be also used to identify the location of specific objects in images. SAM2 was trained on a new dataset called “Segment Anything Video dataset” (SA-V) \cite{ravi2024sam}.

There are two options to run SAM 2: - using an automatic mask generator, that segments all objects in an image or video and generates corresponding masks; - using a point prompt. The second option allows the user to be more specific in object segmentation. To identify the location of an object, the user needs to provide a “prompt”. A prompt can be: a specific point, or series of points, that correspond with the object the user wants to segment; a bounding box that surrounds the object the user wants to segment. If the user provides a specific point as a prompt, he/she can also provide a “negative” prompt that corresponds to a region of an object that he/she does not want to identify. This allows the user to refine his/her prompts from SAM2: if the model predicts a mask that is too big, he/she can remove parts of the mask by specifying a negative prompt.

We use SAM2 to segment and track people in our videos, using as a reference a 2D point on the heads of the people detected by Multi-HMR. This allows us to assign a new ID named IDSAM to each person in each frame.


\begin{figure}[ht!]
\centering
\includegraphics[width=0.1\textwidth]{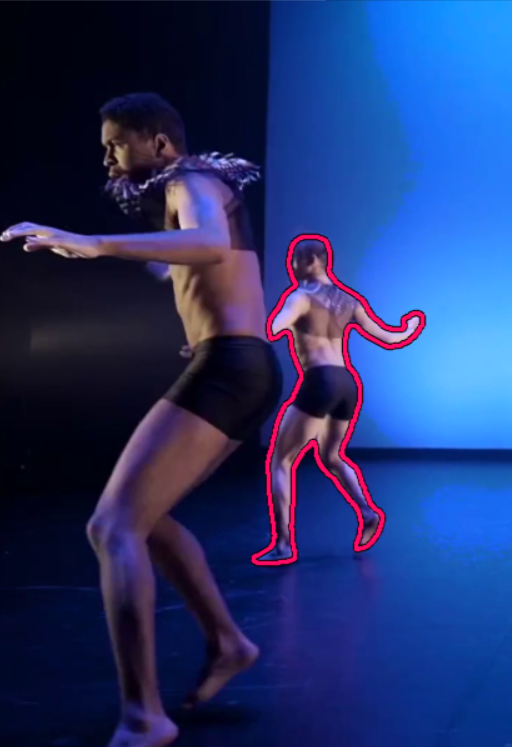}
\includegraphics[width=0.1\textwidth]{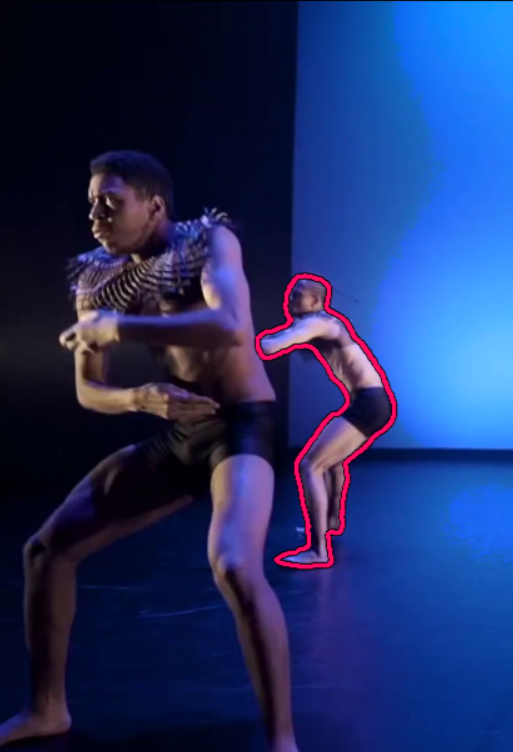}
\includegraphics[width=0.112\textwidth]{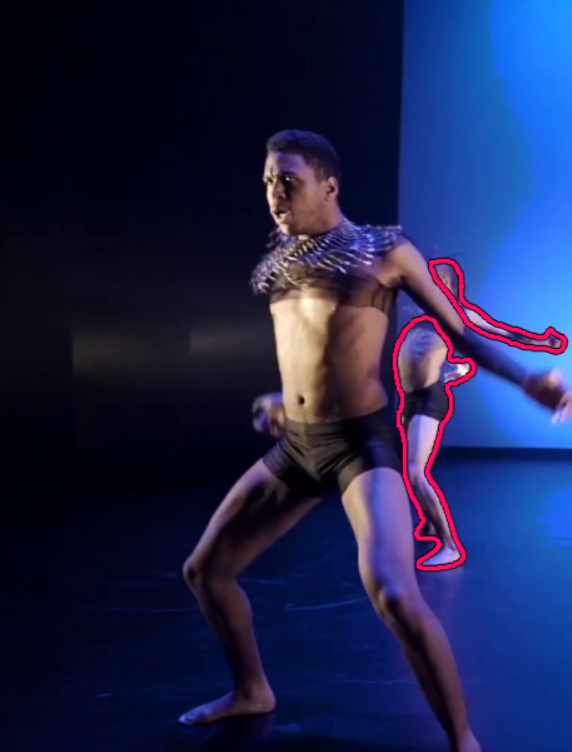}
\includegraphics[width=0.094\textwidth]{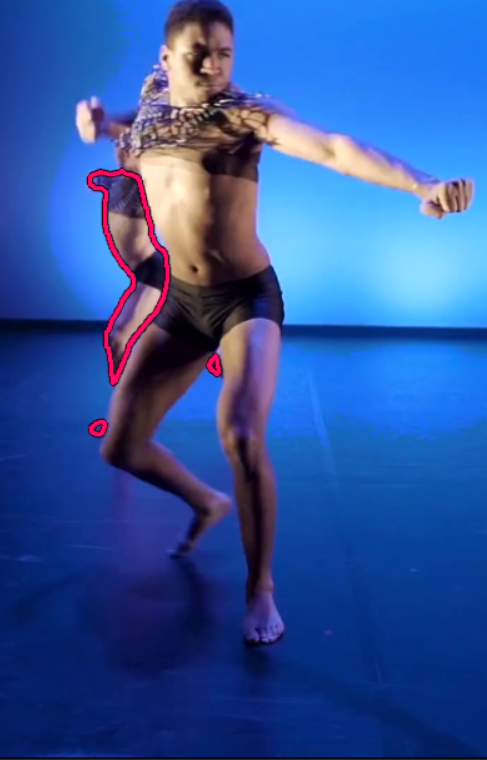}
\caption{(From left to right) Detection of the smaller dancer in the background occluded in some frames by the foreground dancer. The detection is reinforced by the tracking of the segmentation mask through the video sequence. Images from \cite{Talawa2018}.}
    \label{fig17}
\end{figure}

\begin{figure}[ht!]
\centering
\begin{minipage}{0.24\textwidth}
        \centering
        \includegraphics[width=1\textwidth]{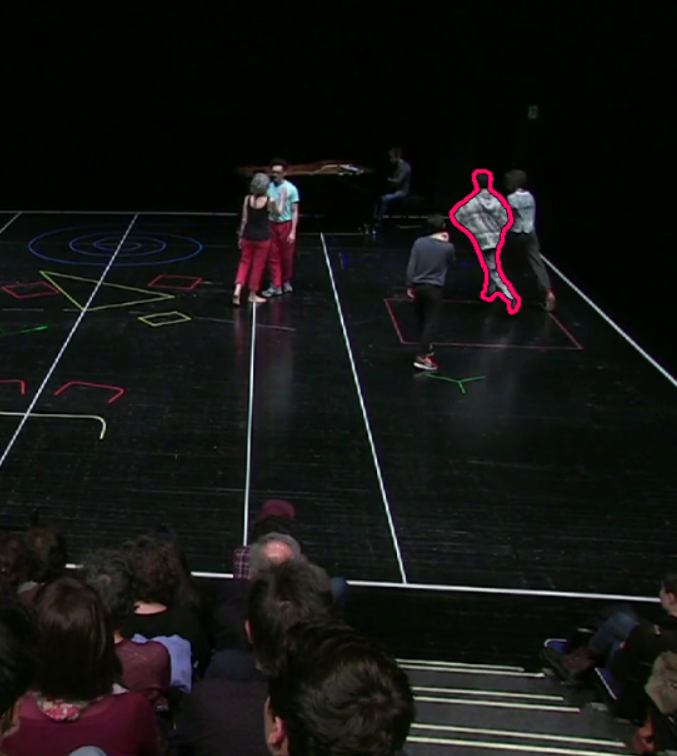}
        \subcaption{}
    \end{minipage}%
    \hspace{0cm}
\begin{minipage}{0.24\textwidth}
        \centering
        \includegraphics[width=1\textwidth]{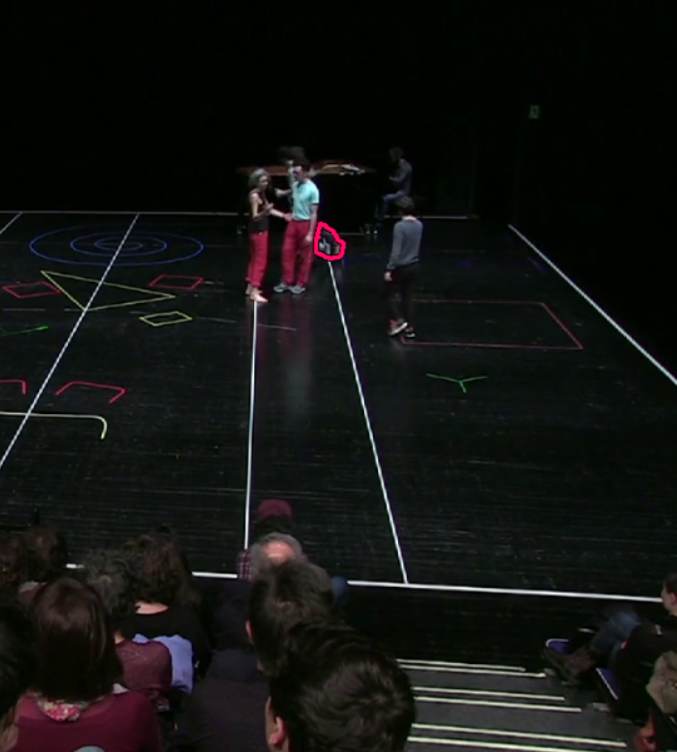}
        \subcaption{}
    \end{minipage}%
    \hspace{0cm}
\begin{minipage}{0.24\textwidth}
        \centering
        \includegraphics[width=1\textwidth]{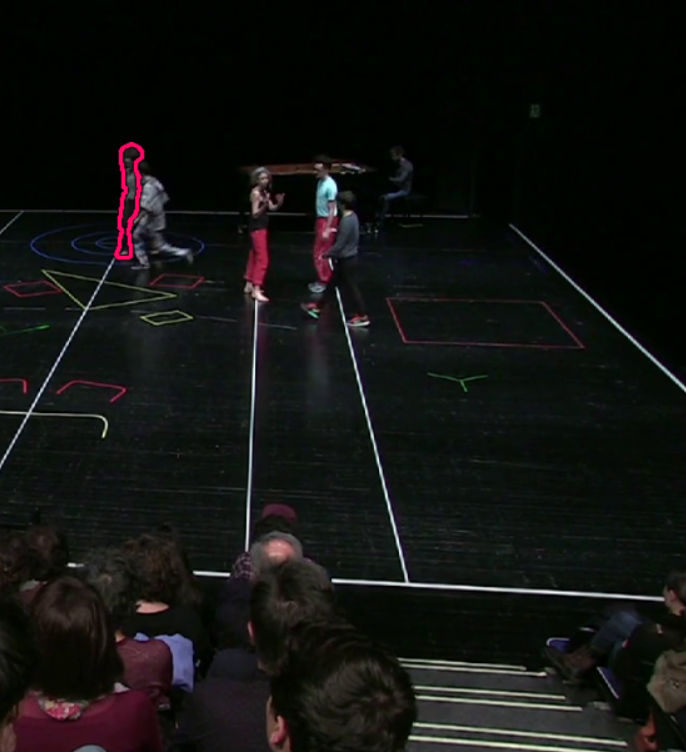}
        \subcaption{}
    \end{minipage}%
    \hspace{0cm}
\caption{Examples of sample frames from a video sequence for which the tracking fails. Images from [D6-Projecto Continuado Continued Project].}
    \label{fig18}
\end{figure}

SAM2 works very well if the number of human pixels tracked is sufficient. On the other hand, it is not suitable when the camera is very far from the scene and the humans are very small. Figure \ref{fig17} shows some examples from the same video where the segmentation of individual frames works well thanks to the tracking. On the other hand, Figure \ref{fig18} shows some examples from another video where the segmentation is wrong.

\subsubsection{Step 7: Re-identification and cleaning} 
The algorithm proposed processes human tracking data from frames stored in a file. First, it identifies the maximum number of humans and tracks their IDs across all frames. Then, it calculates the track size for each human ID, storing tracking data in a structure. For each human, the most frequent IDSAM (the secondary identifier) is analyzed. If the track size exceeds a given threshold (30 frames for a 100 FPS video and 10 frames for a 25 FPS video), the most frequent IDSAM is assigned to the human's ID; otherwise, the ID is set to -1 (this human skeleton is then removed). Combining this 2 information’s allows to make a perfect re-identification on the tracks generated in Step 4.

\begin{figure}[ht!]
    \centering
    \begin{minipage}{0.24\textwidth }
        \centering
        \includegraphics[width=1.1\textwidth,keepaspectratio]{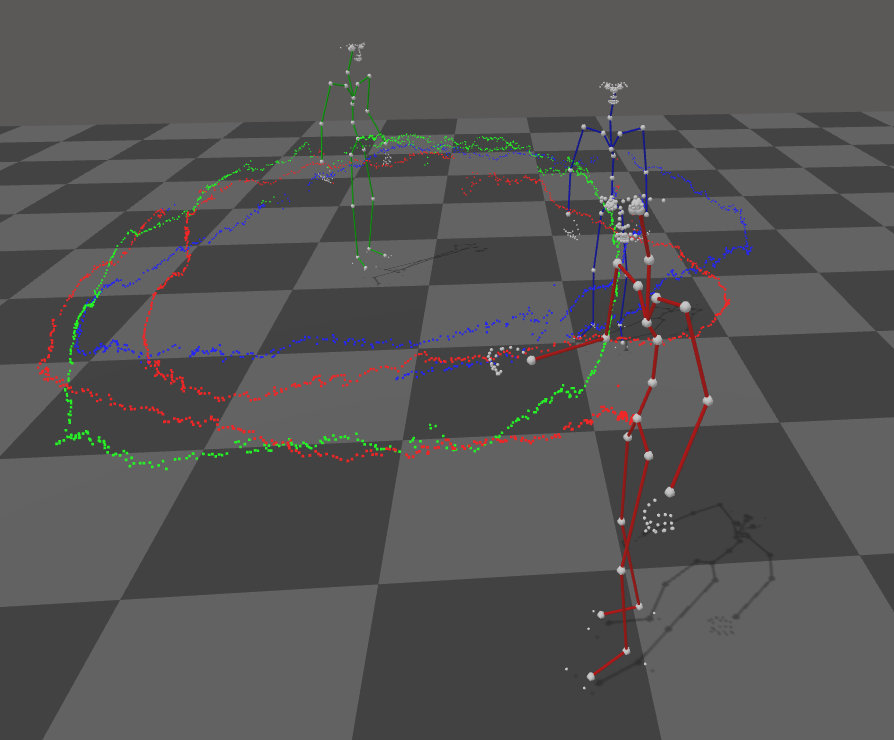}
        \subcaption{}
        \label{fig19.1}
    \end{minipage}%
    \hspace{0cm}
    \begin{minipage}{0.24\textwidth}
        \centering
        \includegraphics[width=0.8\textwidth,keepaspectratio]{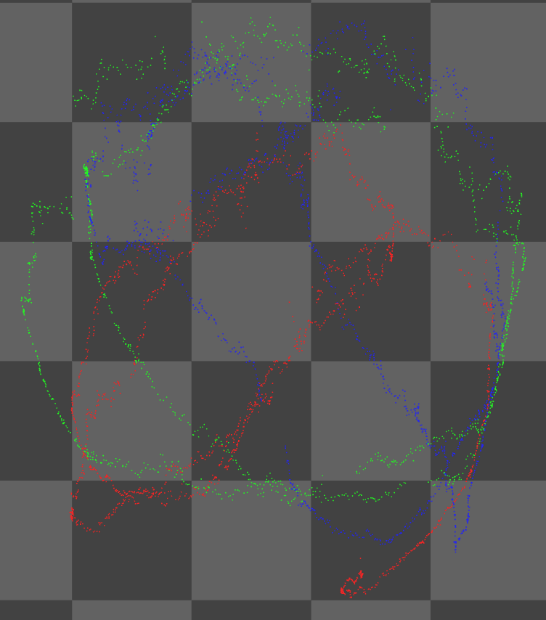}
        \subcaption{}
        \label{fig19.2}
    \end{minipage}
    \caption{ (a) Trajectories and 3D skeletons of re-identified people shown for one point of view, (b) trajectories are shown from above the stage.}
    \label{fig19}
\end{figure}

By merging the data obtained with SAM2 tracking and 3D tracking based on skeleton distances we can eliminate the re-identification errors that can occur when total or partial body occlusions occur.  As illustration, the tracks in Figure \ref{fig13} are perfectly reidentified in Figure \ref{fig19}.

\subsubsection{Step 8: Tracks filtering and interpolation} 
We saw in Step 2 that Multi-HMR's estimation of people's 3D positions is not stable between two consecutive frames, especially for depth estimation. This instability is clearly visible in Figure \ref{fig19}.

\begin{figure}[ht!]
    \centering
    \begin{minipage}{0.24\textwidth }
        \centering
        \includegraphics[width=1.125\textwidth,keepaspectratio]{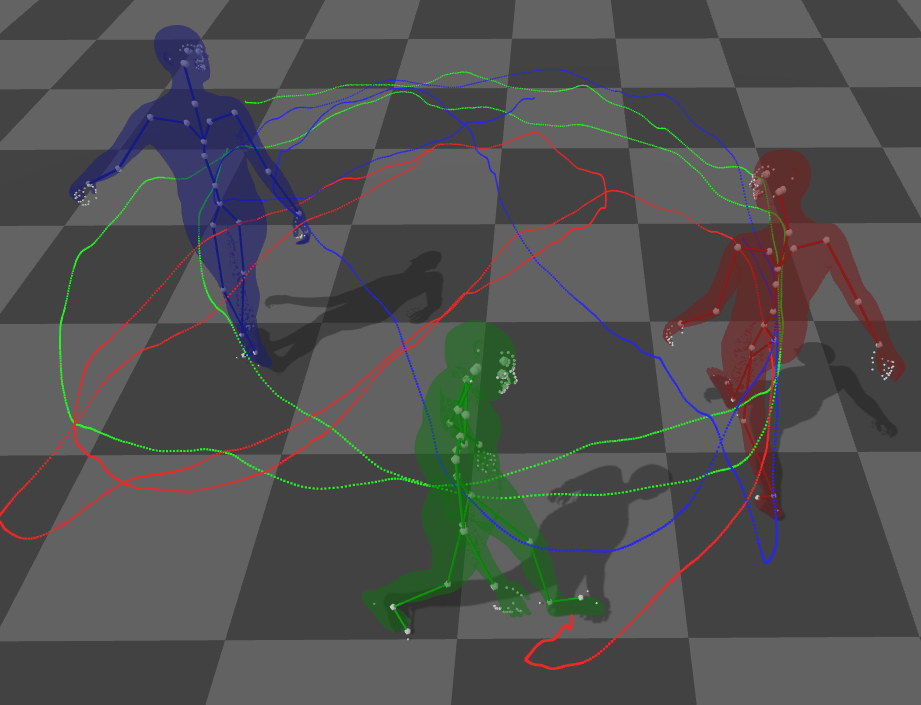}
        \subcaption{}
        \label{fig20.1}
    \end{minipage}%
    \hspace{0cm}
    \begin{minipage}{0.24\textwidth}
        \centering
        \includegraphics[width=0.75\textwidth,keepaspectratio]{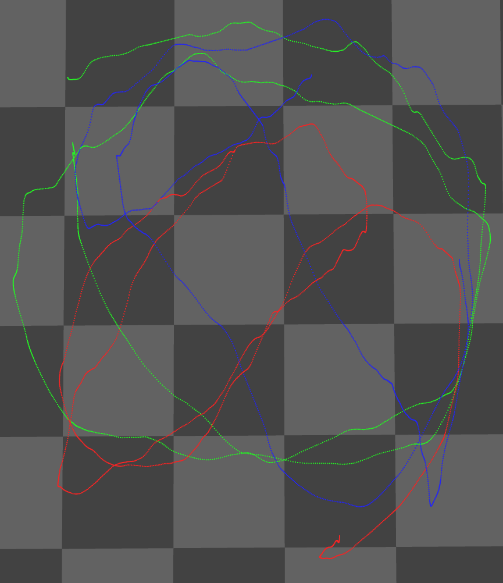}
        \subcaption{}
        \label{fig20.2}
    \end{minipage}
    \caption{ (a) Filtered and interpolated trajectories, and 3D skeletons/meshes shown for one point of view, (b) trajectories shown from above the stage.}
    \label{fig20}
\end{figure}

Multi-HMR is also unable to handle occlusions, meanwhile the re-identification method detailed in Step 7 can also cause people to disappear in certain frames. To overcome these problems, we chose to use a Radial Basis Function (RBF) interpolator with a linear kernel. Radial Basis Function (RBF) interpolation is a method for interpolating scattered data in multiple dimensions \cite{baxter2010}. It is particularly useful when the data points are irregularly spaced (which is our case because of the occlusions). The basic idea is to represent the interpolated surface as a weighted sum of radial basis functions centered at the data points. This numerical method can be used for both interpolation and approximation, depending on the parameters provided. For our case study we choose to use a method that constructs a function that approximates the given data points (to smooth trajectories) and add people when they are missing. This is achieved by introducing a smoothing parameter that allows the function to deviate from the exact data points to achieve a smoother result (see Figure \ref{fig20}). The value of this smoothing parameter has been defined empirically from tests and experiments done with dance video archives.
\subsubsection{Step 9: Estimation of camera motion and 3D reconstruction}
\begin{figure}[ht!]
\centering
\includegraphics[width=0.4\textwidth]{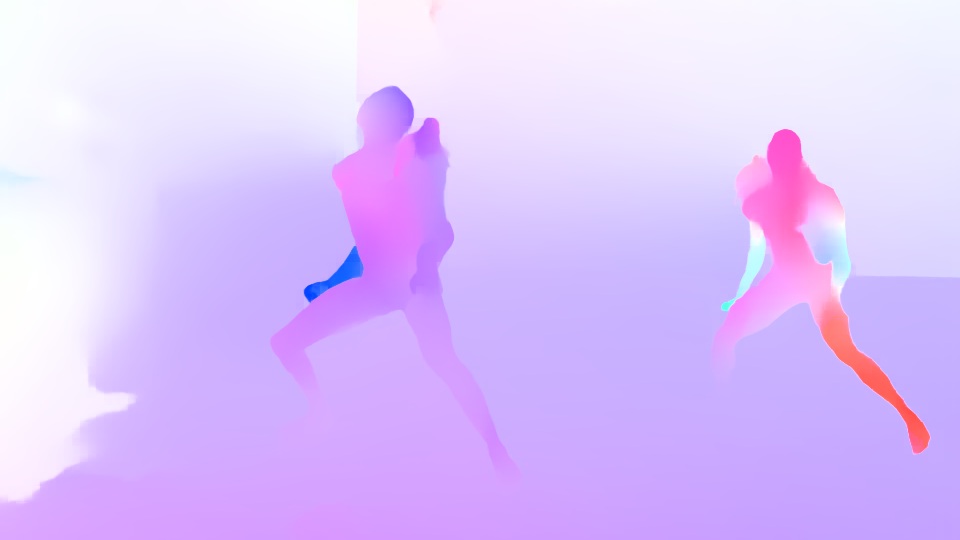}
\caption{Optical flow from two consecutive frames. The optical flow map highlights the camera motion over time (flow vectors are color-coded in this image). Images from  \cite{Talawa2018}.}
    \label{fig21}
\end{figure}

Camera motion refers to the movement or changes in position, orientation, or perspective of the camera while capturing the video (as example see Figure \ref{fig21}). It can include camera pans, tilts, zooms, rotations, or any other camera movements. Camera motion analysis is important in video processing to understand and account for camera dynamics, stabilize footage, or detect specific motion patterns within the scene.

\begin{figure}[ht!]
    \centering
    \begin{minipage}{0.16\textwidth}
        \centering
        \includegraphics[width=\textwidth]{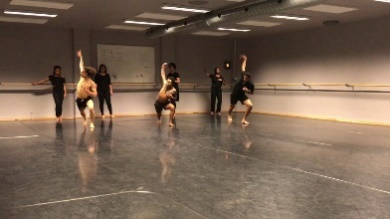}
        \subcaption{}
        \label{fig22.1}
    \end{minipage}%
    \hfill
    \begin{minipage}{0.16\textwidth}
        \centering
        \includegraphics[width=\textwidth]{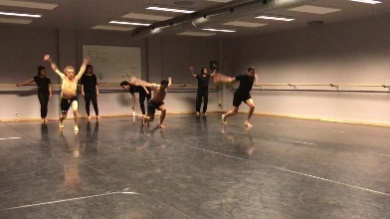}
        \subcaption{}
        \label{fig22.2}
    \end{minipage}%
    \hfill
    \begin{minipage}{0.16\textwidth}
        \centering
        \includegraphics[width=\textwidth]{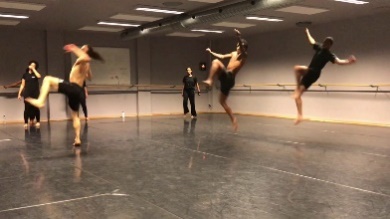}
        \subcaption{}
        \label{fig22.3}
    \end{minipage}
    \vskip\baselineskip
     \begin{minipage}{0.22\textwidth}
        \centering
        \includegraphics[width=\textwidth]{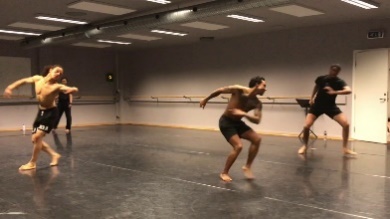}
        \subcaption{}
        \label{fig22.4}
    \end{minipage}
    \hspace{0cm}
    \begin{minipage}{0.22\textwidth}
        \centering
        \includegraphics[width=\textwidth]{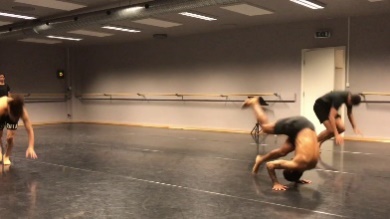}
        \subcaption{}
        \label{fig22.5}
    \end{minipage}
    \vskip\baselineskip
    \begin{minipage}{0.35\textwidth}
        \centering
        \includegraphics[width=\textwidth]{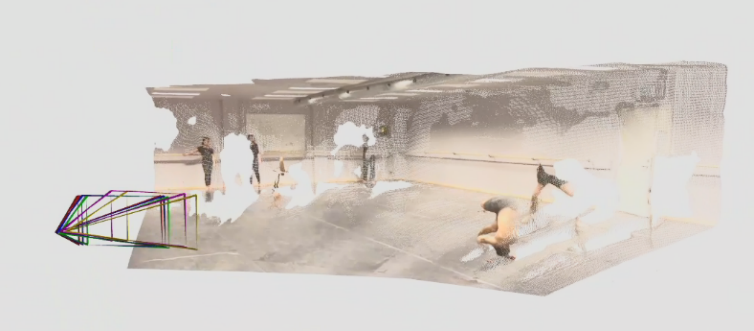}
        \subcaption{}
        \label{fig22.6}
    \end{minipage}
    \caption{(a) Frame N°1, (b) frame N+1, (c) frame 2N+1, (d) frame 3N+1, (e) etc. (f) 3D reconstruction of the scene in the world coordinate system with camera position values from one frame to another one. In this sequence we have a horizontal motion of the camera. Images from \cite{Talawa2018}.}
    \label{fig22}
\end{figure}

To estimate the motion of a camera and to perform 3D reconstruction from 2D data, we used MASt3R – Matching And Stereo 3D Reconstruction \cite{leroy2024}. MASt3R is a 3D reconstruction model capable of handling thousands of images. MASt3R brings higher precision and detail to 3D reconstruction and localization tasks by providing pixel correspondences for even very large image collections. MASt3R is based on the previous DUSt3R framework and a matching algorithm. It efficiently outputs a metric 3D reconstruction along with dense local feature maps, providing accurate depth perception and spatial understanding (see Figure  \ref{fig22}).

The presence of people dancing generally causes problems for MASt3R's camera pose estimation. However, the in-depth study of MASt3R's behaviour that we performed showed that such movement does not significantly affect camera pose estimation, especially for dance video sequences with small camera movements.

%% file: text/Results.tex
\subsection{Visualization tools}
Each stage of the pipeline proposed was implemented in the form as a Python script with a command line that  accepts the previously outlined input/output parameters. Below, we detail several visual analysis tools developed to analyze the results of this pipeline.

\subsubsection{Data visualization tool}
\begin{figure}[ht!]
    \centering
    \includegraphics[width=0.45\textwidth]{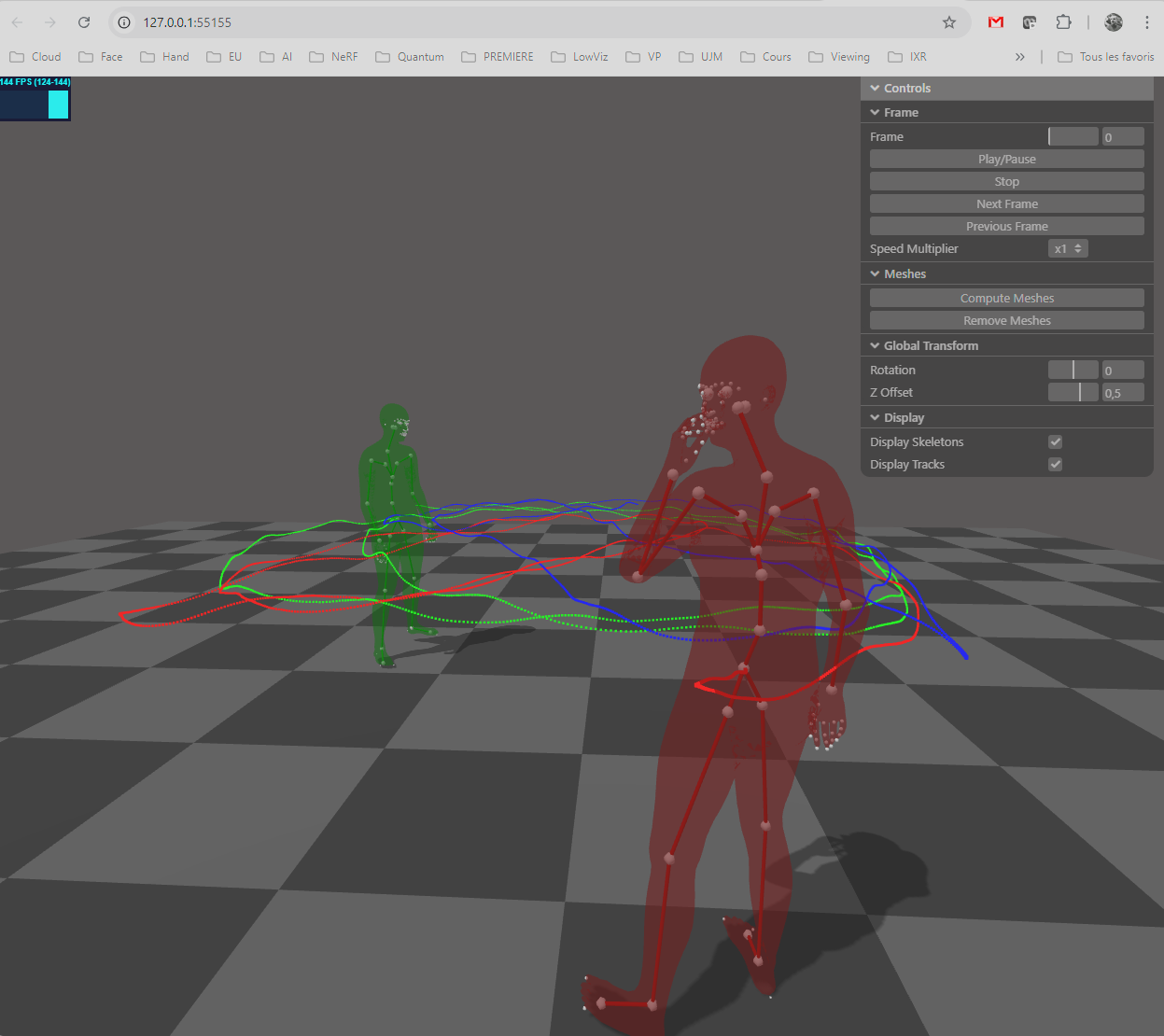}
    \caption{Segmentation mask (with or without the skeleton) of each person detected in a video sequence, with or without trajectories of people in the 3D space extracted from this video sequence. The viewing position of this 3D reconstruction can be modified by the user. The interface enables the user to select several visualization modes listed in the control panel on the upper right.}
    \label{fig23}
\end{figure}

The files generated by our pipeline can all be visualized using a data visualization tool created specifically for this pipeline. This tool is based on the Flask module in python (for the backend) and HTML/JavaScript with Three.js associated with a web browser (for the frontend), see Figure \ref{fig23}.
It allows to:

\begin{itemize}
    \item View all poses in a given frame in the form of a skeleton or 3D body shape (this 3D shape is only visible on demand, as it has to be recalculated by the Python back-end via Flask).
    \item View all skeleton tracks (associated with a color corresponding to the track ID).
    \item Animate skeletons at different speeds.
\end{itemize}

\subsubsection{XR visualization tool} 
\begin{figure}[ht!]
    \centering
    \includegraphics[width=0.45\textwidth]{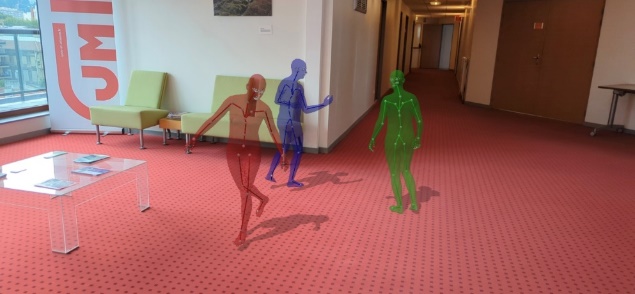}
    \caption{Visualization of the skeleton and of 3D shape of three dancers in an augmented environment.}
     \label{fig24}
\end{figure}

Additionally, we have developed a web visualization tool that is entirely independent of Flask and can be integrated with the relevant data on a static website. This tool leverages existing code and incorporates two new WebXR-based visualization modes: an eXtended Reality (XR) mode that creates an augmented reality (AR) experience with Android smartphones and tablets (WebXR is not yet supported on iPhone and iPad). Additionally, the tool offers a mixed reality (MR) experience with Meta Quest 2, 3, and Pro headsets (it is also compatible with Apple's Vision Pro, though this has not yet been tested). It also features a virtual reality (VR) mode, which has been tested on Meta Quest 2, 3, and Pro. As example, see Figure  \ref{fig24}.

\begin{figure}[ht!]
    \centering
    \includegraphics[width=0.45\textwidth]{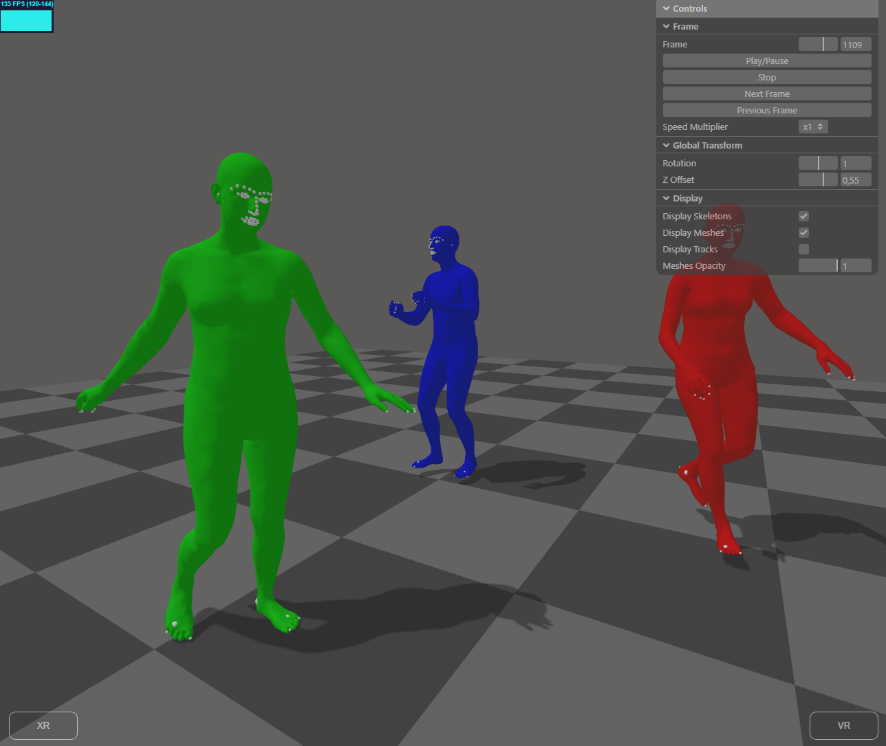}
    \caption{3D mesh of each person detected in a video sequence. The interface enables also to display on VR display: skeletons of each dancer, trajectories on the stage along the video sequence, shadows on the ground.}
     \label{fig25}
\end{figure}

For each video sequence processed, the following data can be visualized (as example see Figure \ref{fig25}):

\begin{itemize}
    \item Skeletons of bodies within a frame.
    \item The 3D shapes of the bodies within a frame, represented as a set of triangles for each body, defined by vertices and their associated normals.
    \item The tracks modelized by a set of points with colors defined by their IDs.
\end{itemize}

A Python script was developed to generate two files (one JSON and one binary) containing all the information required for this tool:
\begin{itemize}
    \item All the tracks (in the JSON file).
    \item Skeletons, vertices, and body normals for each frame (in the binary file).
\end{itemize}
The data generated by this script can be large. For example, a four-minute scene with two dancers obtained from a 100-frame-per-second video generates a 15 GB binary file.

Due to the size of these files, they cannot be read by a web browser. Instead, they are streamed into our visualization tool. To ensure fluid visualization, it is necessary for the web server to be on the same local network as the web client (which in this case is the web browser in charge for the visualization).


More results are shown on this website: \url{https://www.couleur.org/articles/arXiv-1-2025/}

%% file: text/Discussion.tex
In this paper we proposed a video processing pipeline to extract from dance archive videos: - the separate sequences in a video; - low level features (camera position, 2D and 3D pose of human body skeletons, bounding box and 3D meshes of human bodies); - high level features (human body positions, human body trajectories, human body tracking).

In our developments, we also tried to estimate the light sources position and direction but the preliminary results we get were not satisfying. We
think that ligth position and direction estimation could help in some scenarios to improve human body pose estimation. In a future work we will study how to accurately estimate lighting conditions from archive videos, including changes of lighting conditions in a video sequence, and how to relight video contents based on lighting estimation.

\begin{figure}[ht!]
    \centering
    \begin{minipage}{0.22\textwidth}
        \centering
        \includegraphics[width=\textwidth, keepaspectratio ]{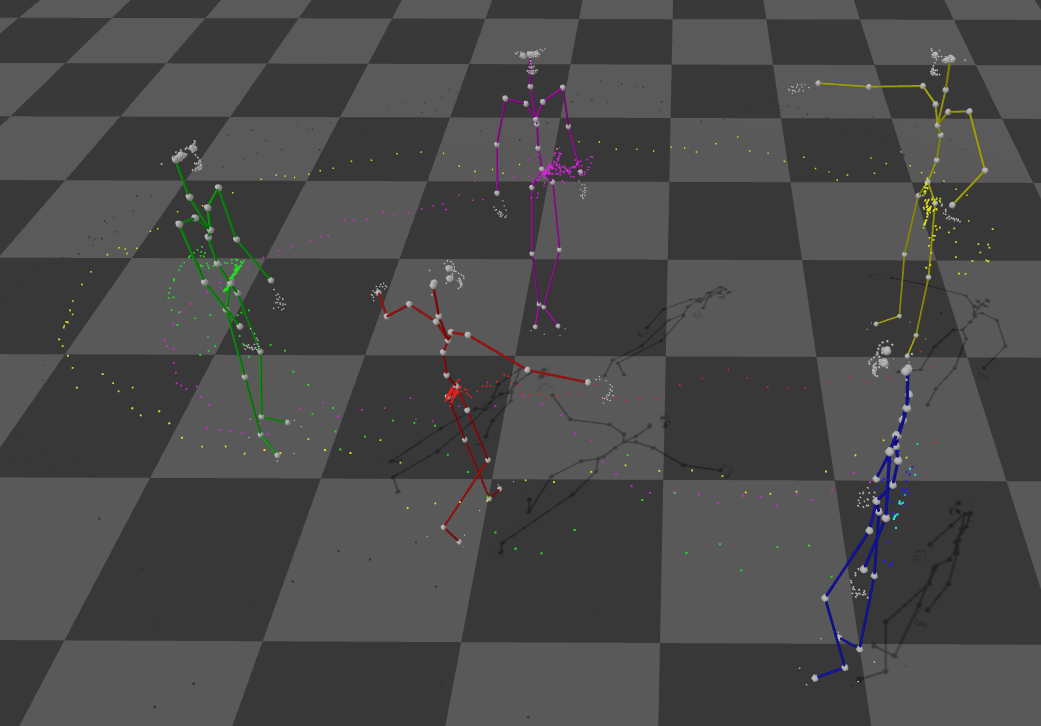}
        \subcaption{}
        \label{fig26.1}
    \end{minipage}%
    \hspace{0cm}
    \begin{minipage}{0.22\textwidth}
        \centering
        \includegraphics[width=\textwidth, keepaspectratio]{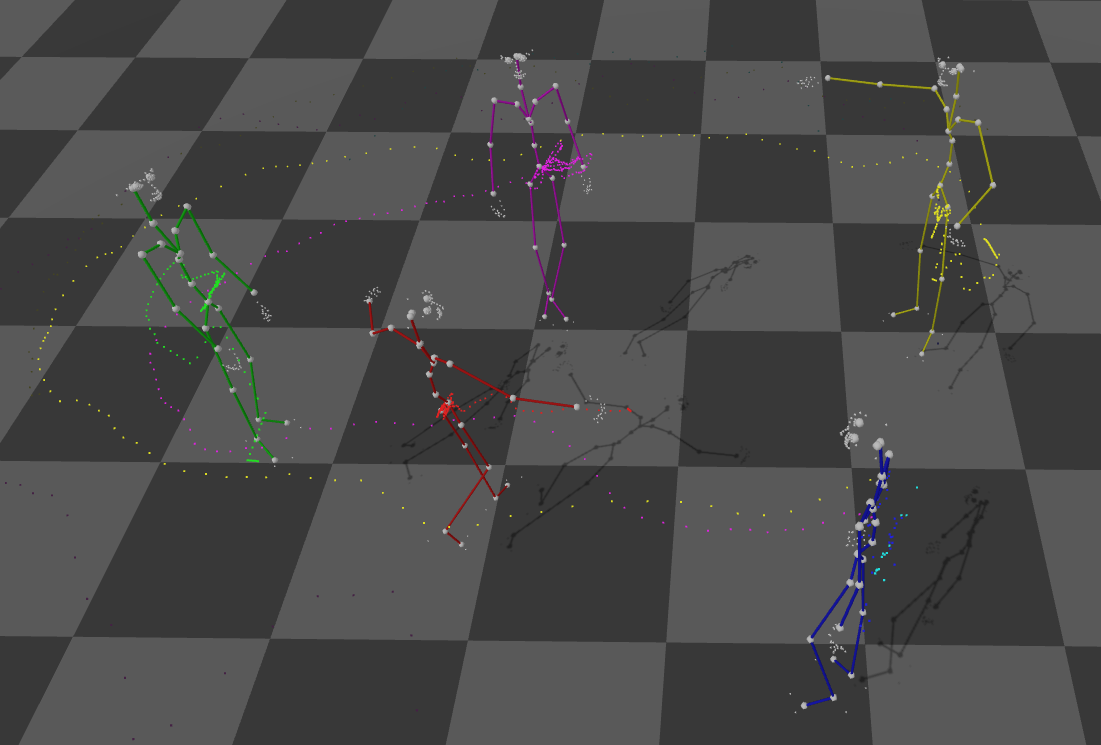}
        \subcaption{}
        \label{fig26.2}
    \end{minipage}
   \vspace{-0.1cm}
    \begin{minipage}{0.22\textwidth}
        \centering
        \includegraphics[width=\textwidth, keepaspectratio]{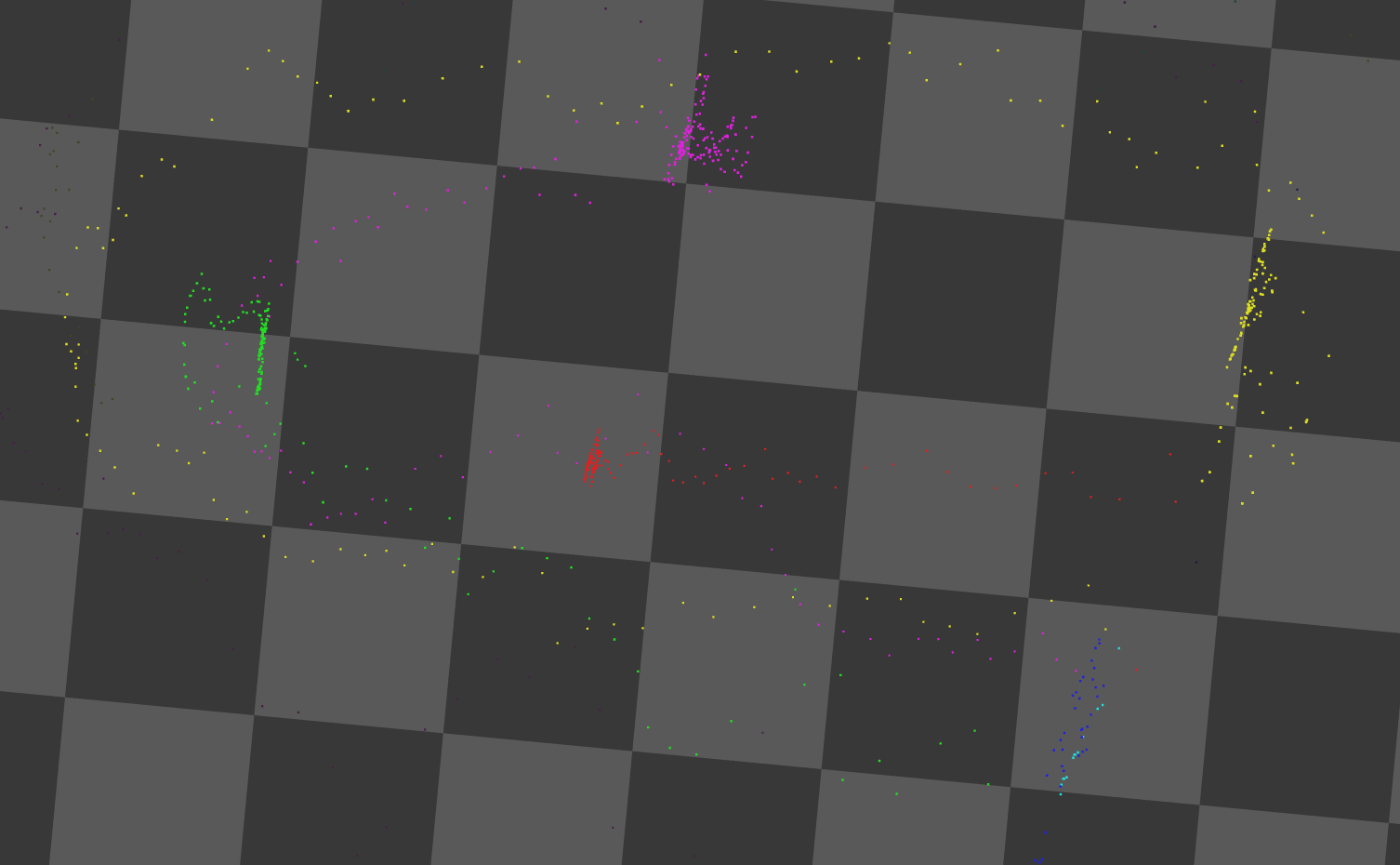}
        \subcaption{}
        \label{fig26.3}
    \end{minipage}
    \hspace{0cm}
    \begin{minipage}{0.22\textwidth}
        \centering
        \includegraphics[width=\textwidth, keepaspectratio]{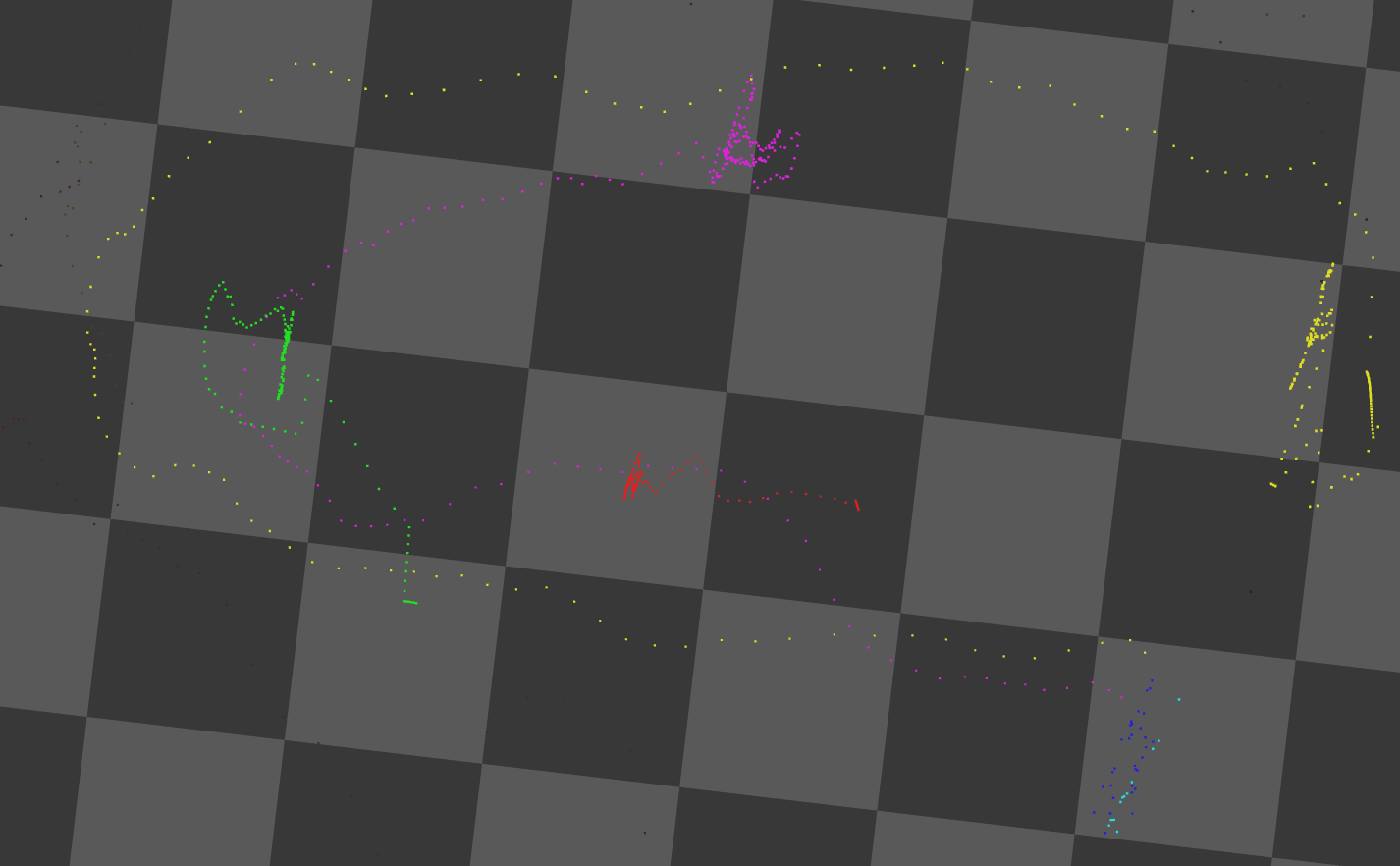}
        \subcaption{}
        \label{fig26.4}
    \end{minipage}
    \caption{Examples of result of tracks before and after applying interpolation on sequence from [Projecto Continuado Continued Project]. (a) Noisy estimation with skeleton and track. (b) Smooth estimation with skeleton after filtering with track (c) Noisy estimation of track (top view). (d) Smooth estimation of track after applying filtering (top view).}
    \label{fig26}
\end{figure}

\begin{figure}[ht!]
    \centering
    \begin{minipage}{0.45\textwidth}
        \centering
        \includegraphics[width=\textwidth]{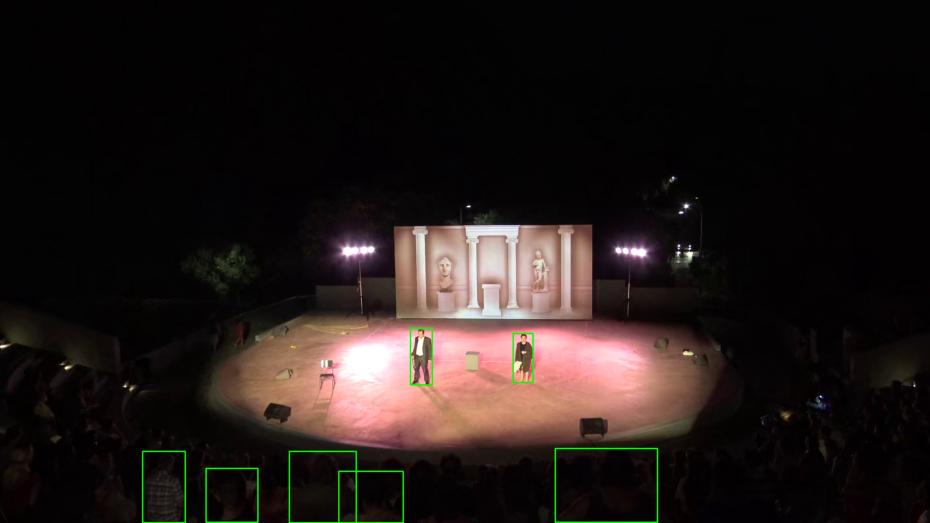}
        \subcaption{}
        \label{fig27.1}
    \end{minipage}
    \hspace{0cm}
    \begin{minipage}{0.45\textwidth}
        \centering
        \includegraphics[width=\textwidth]{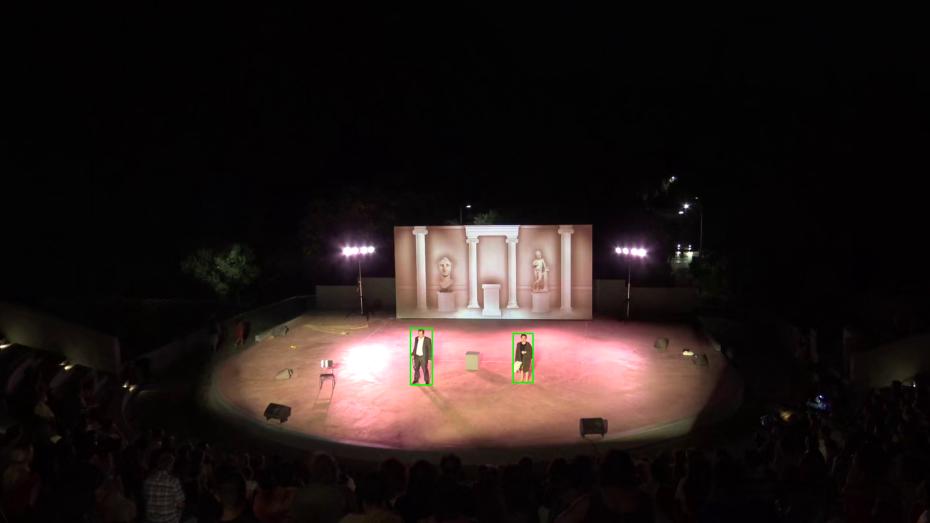}
        \subcaption{}
        \label{fig27.2}
    \end{minipage}
    \caption{(a) Public poses a challenge to the model: it can not differentiate dancers from non-dancers. (b). A filtering technique based on the Region Of Interest (ROI) mechanism can remove detections of people in the audiaence and produce refined results.}
     \label{fig27}
\end{figure}

\begin{figure}[ht!]
    \centering
    \begin{minipage}{0.23\textwidth }
        \centering
         \includegraphics[width=\textwidth]{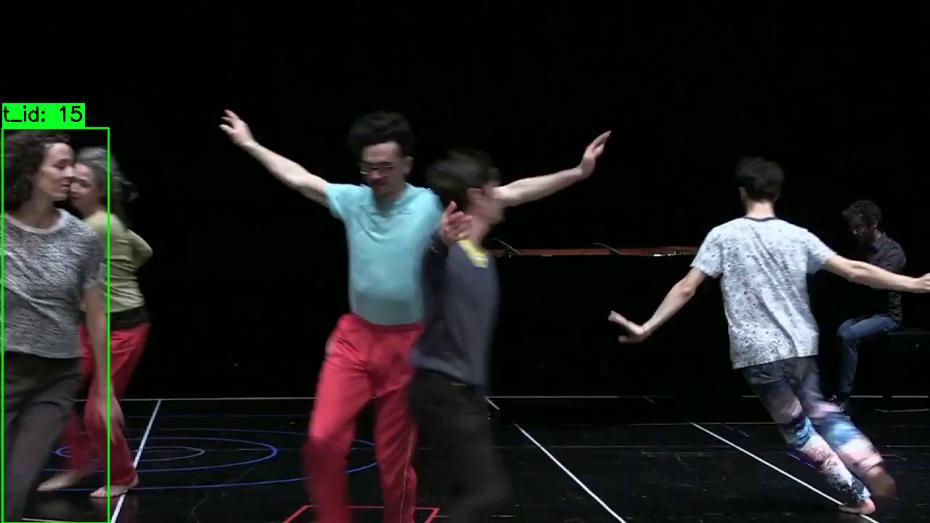}
        \subcaption{}
        \label{fig28.1}
    \end{minipage}%
    \hspace{0cm}
    \begin{minipage}{0.23\textwidth}
        \centering
         \includegraphics[width=\textwidth]{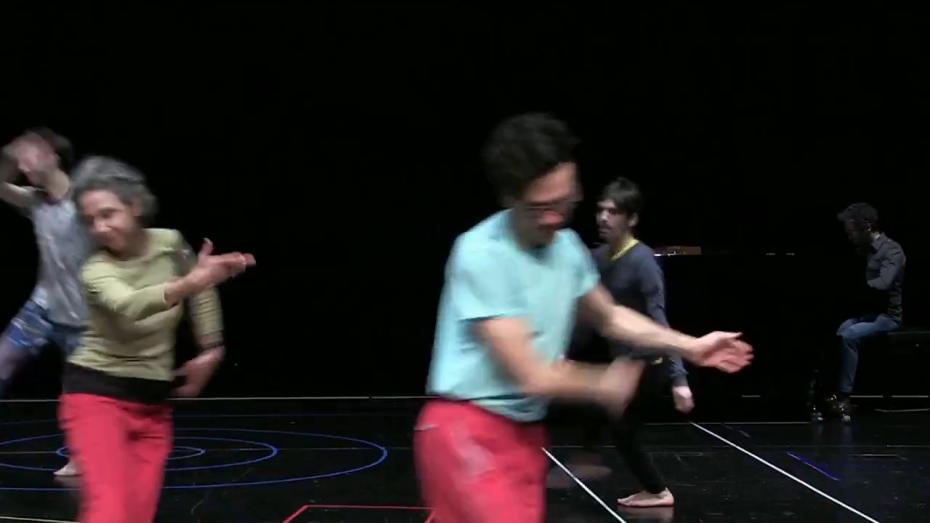}
        \subcaption{}
        \label{fig28.2}
    \end{minipage}
    \hspace{0cm}
    \begin{minipage}{0.23\textwidth}
        \centering
        \includegraphics[width=\textwidth]{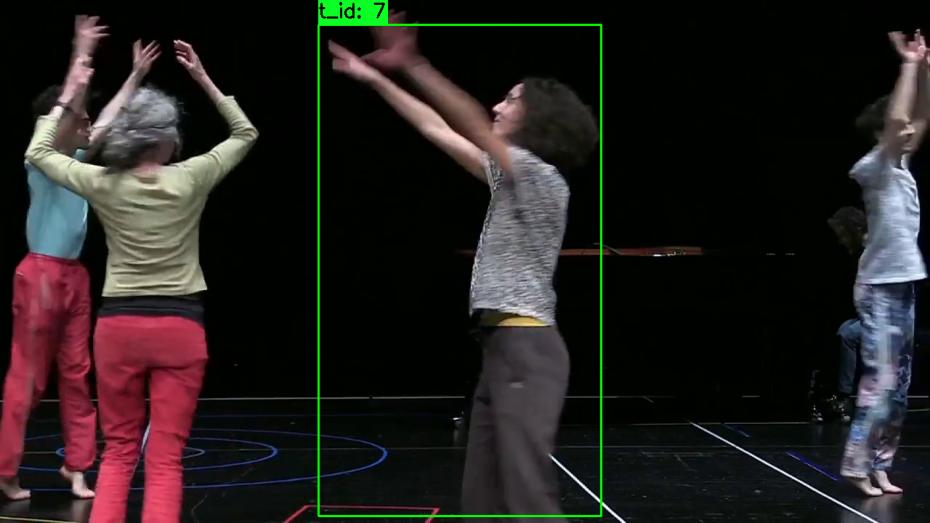}
        \subcaption{}
        \label{fig28.3}
    \end{minipage}
    \caption{(a) After detection the people on the left was assigned the tracking id (t\_id=15). (b) This person disappears from the visual field. (c) This person re-appears on the stage and the tracker assigned her a new tracking id (t\_id=7). Even if the person is the same the tracker is unable to re-identify her again after re-appearing. This gives raise to a common challenge in multi-object tracking called re-identification i.e., tracker being unable to re-identify the same object or person.}
    \label{fig28}
\end{figure}

Section 2 provided an overview of the different computer vision tasks that have been implemented and the computer vision methods that have been used. In order to limit inaccurate detections, missing detections, or noisy estimations, few technical steps have been included to this pipeline, especially in steps 3 and 8 (as example see (see Figure \ref{fig26}). Other filtering steps could be added to the pipeline, for example to distinguish the performers from the public (see Figure \ref{fig27}), or to re-identify people coming back on the stage after moving out (see Figure \ref{fig28}). The current pipeline was developed to detect and track human bodies in archive videos without public in the camera field, without long occlusions and intertwining between human bodies.

The accuracy and robustness of human pose estimation models is a hot topic in general, especially in dance as the precision of gestures and movements is very important in dance. The main problem of existing pose estimation models is that they have not being trained on a variety of dance datasets nor on atypical movements or postures specific to dance. The main objective of our investigations was to strengthen the robustness of pose estimation against occlusion, shadows and varying lighting conditions. We did extensive tests and experiments from various PREMIERE archive videos and from the PREMIERE Dance Motion Dataset specifically designed for this purpose. Several examples of results are available on this website \url{https://www.couleur.org/articles/arXiv-1-2025/}

Due to a lack of dance datasets with ground truth information, especially in regards to depth information, we did not yet evaluate and compare quantitatively our pose estimation pipeline with other pose estimation methods. We only perform visual evaluation and comparison using visual analytic tools. In a future work we will perform quantitative evaluations using the DanceTrack dataset \cite{dancetrack2022}. This dataset contains a collection of different dance videos along with their annotations. Specifically, it contains 100 videos divided in three subsets: training (40 videos), validation (25 videos) and test (35 videos). However, only the training and validation subsets are made public. Furthermore, the provided annotations include the bounding box and identification information. Owing to that, we aim to use validation dataset to evaluate and compare our pipeline with existing methods particularly for detection and tracking tasks. We will also investigate accurate depth estimation methods to improve the accuracy of 3D pose estimation, especially in regards to depth estimation.

We demonstrate in our work that it is very challenging of a user to select the most appropriate pose estimation method for a given video content, as the efficiency and relevance of a method depends of: the video content (such as the speed of movements, the occlusions); the quality of the video (such as the spatial resolution, the contrast); and of the setting values used (for the parameters). To face this issue we propose a user interface that enables a user to refine the pose estimation results based on his/her preferences through the use of visual analytical tools. Several examples of results are shown in the webpage \url{https://www.couleur.org/articles/arXiv-1-2025/}. With the rapid progresses in pose estimation based on artificial intelligence we are confident that in few months we will be able to propose an automatic pose detection method for dance analysis.